\documentclass[11pt]{article}

\usepackage[preprint]{acl}

\usepackage{times}
\usepackage{latexsym}

\usepackage[T1]{fontenc}

\usepackage[utf8]{inputenc}

\usepackage{microtype}

\usepackage{inconsolata}

\usepackage{graphicx}

\usepackage{amsmath}
\usepackage{amssymb}
\usepackage{wrapfig}
\usepackage{multirow}
\usepackage{booktabs}
\usepackage{enumitem}
\usepackage{algorithm}
\usepackage{algpseudocode}
\usepackage{amsmath}
\usepackage{colortbl} 
\usepackage{xcolor}   
\usepackage{pifont}
\newcommand{\cmark}{\textcolor{green!60!black}{\ding{51}}}
\newcommand{\xmark}{\textcolor{red!70!black}{\ding{55}}}
\usepackage[most]{tcolorbox}

\newtcolorbox{questionbox}{
  colback=gray!5,
  colframe=gray!45,
  boxrule=0.4pt,
  arc=1mm,
  left=4pt,
  right=4pt,
  top=4pt,
  bottom=4pt,
  fonttitle=\bfseries,
  title=Ablation Questions
}

\title{Make LLM Learn to Synthesize from Streaming  \\  Experiences through Feedback}

\author{
 \textbf{Zhenlin Hu\textsuperscript{1\text{*}}},
 \textbf{Yan Wang\textsuperscript{2\text{*}}},
 \textbf{Zhen Bi\textsuperscript{1,4\text{\dag}}},
 \textbf{Zihao Xue\textsuperscript{1}},
 \textbf{Bingyu Zhu\textsuperscript{2}},
\\
 \textbf{Longtao Huang\textsuperscript{2}},
 \textbf{Xiongtao Zhang\textsuperscript{1,4}},
 \textbf{Zeyu Yang\textsuperscript{1}},
 \textbf{Zhixuan Chu\textsuperscript{3}},
 \textbf{Jungang Lou\textsuperscript{1,4\text{\dag}}}
\\
 \textsuperscript{1}Huzhou Normal University,
  \textsuperscript{2}Alibaba Group,
  \textsuperscript{3}Zhejiang University,
\\
  \textsuperscript{4}Zhejiang Key Laboratory of Intelligent Education Technology and Application
\\
\text{\textsuperscript{*}These authors contributed equally to this work}
\text{\textsuperscript{\dag}Corresponding authors}
}

\begin{document}
\maketitle
\begin{abstract}
Large language models (LLMs) have been widely adopted for synthetic data generation, significantly reducing annotation costs. 
However, most existing studies treat synthesis as a set of isolated tasks and overlook a more fundamental question: whether a model can \emph{learn to synthesize} by accumulating experience from past tasks and transferring it to future ones.
In this work, we introduce \textbf{StreamSynth}, a new setting in which synthesis tasks arrive sequentially and experience from historical tasks provides informative signals for future synthesis. 
To address this setting, we propose \textbf{SynLearner}, a general framework that enables synthesis models to acquire reusable synthesis experience over a task stream. 
Instead of generating data independently for each task, SynLearner encourages the model to explore diverse synthesis patterns, learn from feedback, and balance sample quality with set-level diversity as tasks evolve.
Extensive experiments across multiple benchmarks show that SynLearner effectively leverages experience from earlier tasks to improve synthesis performance on later ones, exhibiting consistent cross-task transferability. 
These findings provide evidence for the feasibility of \textbf{StreamSynth} and highlight synthetic data generation as an experience-driven process that can benefit from task streams.
\end{abstract}

\section{Introduction}

Large language models (LLMs) have been widely adopted for synthetic data construction \citep{TinyStories,Self-Instruct,jas_ours, 37_xue_2026quitobench}.
Prior work has used LLMs for annotation generation, data augmentation, and few-shot dataset construction, substantially reducing manual labeling costs while expanding task coverage \citep{MetaSynth,Automatic_Instruction,synthetic_data_1, 38_xue_2026coreb}. Despite these advances, most existing approaches focus on improving generation quality within single or isolated tasks \citep{synthetic_prompt_1,MathSmith}. A more fundamental question remains largely unexplored: how can a model \emph{learn to synthesize}, namely extract transferable synthesis strategies from past experience and generalize them to future tasks?

\begin{figure}[t!] 
  \hspace*{\fill} 
  \includegraphics[width=1.0\linewidth]{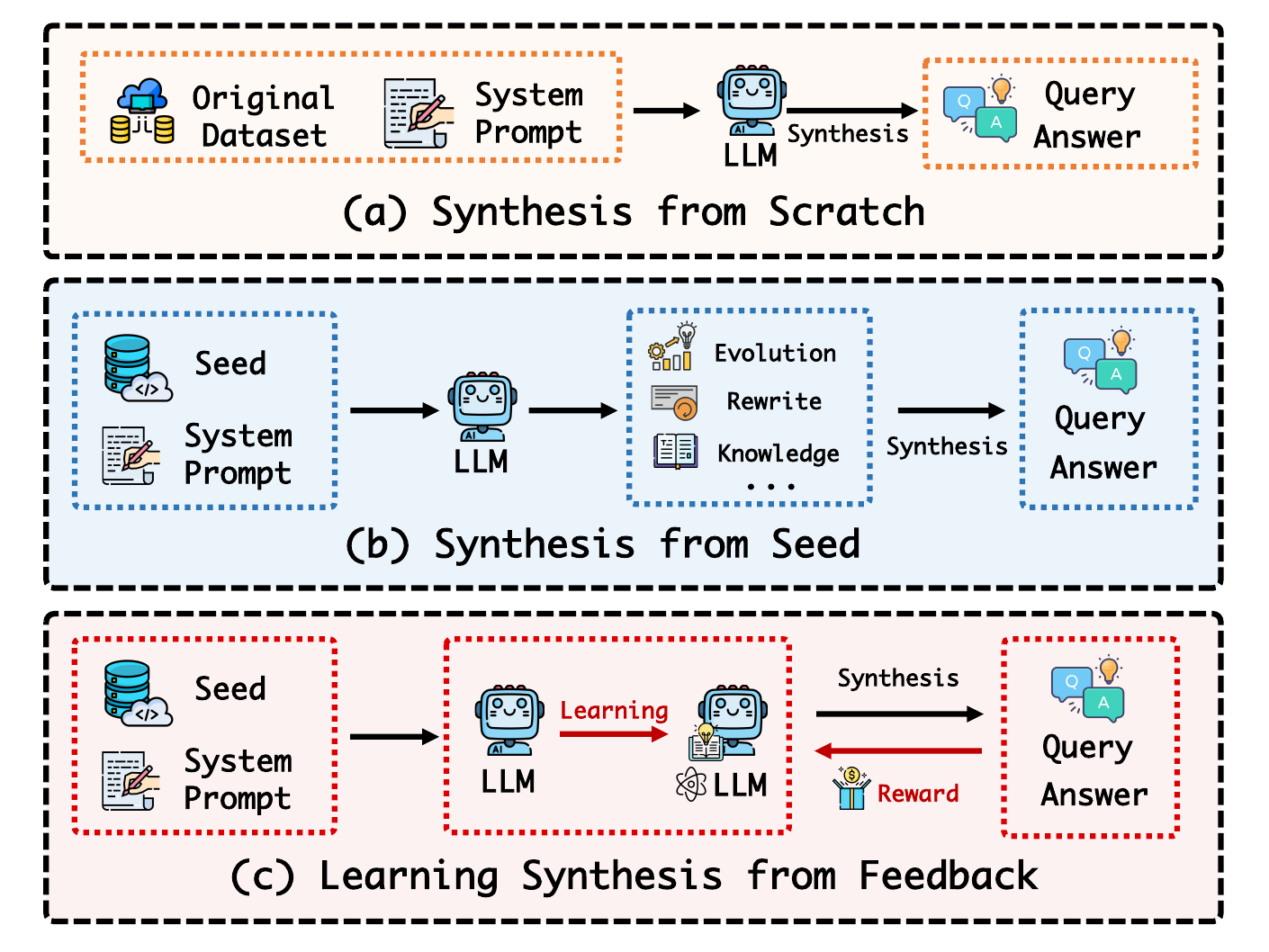} 
  \caption{Comparison of synthesis paradigms.
(a) \textit{From Scratch}: generation guided by fixed rules or heuristics.
(b) \textit{From Seed}: expansion from small seed sets.
(c) \textit{Learning from Feedback}: our proposed paradigm where the synthesis model progressively improves its synthesis behavior through feedback and prior experience.
}
  \label{fig:motivation_fig}
\end{figure}
Existing synthesis approaches can be broadly categorized into two paradigms, as summarized in Figure~\ref{fig:motivation_fig}. 
The first is \textit{synthesis from scratch}, where LLMs generate data directly from task descriptions, prompts, or heuristic rules (Figure~\ref{fig:motivation_fig}a). 
The second is \textit{synthesis from seed}, where seed examples are expanded through rewriting, evolution, or knowledge augmentation (Figure~\ref{fig:motivation_fig}b). 
Although effective, both paradigms usually treat synthesis as a static, task-specific procedure, rather than a capability that can be accumulated and transferred across tasks.
In contrast, we advocate \textit{learning to synthesize from feedback}, where the synthesis model refines its generation behavior through feedback signals and accumulated experience (Figure~\ref{fig:motivation_fig}c). 
This motivates our proposed \textbf{StreamSynth} setting, which studies synthesis as an experience-driven learning process over task streams.


Motivated by this perspective, we formalize a new task setting termed \textbf{StreamSynth} (Streaming Synthesis). 
In StreamSynth, synthesis tasks arrive sequentially as a stream, and the goal is to train a synthesis model that can acquire, reuse, and generalize synthesis experience across tasks. 
Unlike conventional approaches that treat each task independently or repeatedly rely on generic large language models \citep{GPT-4,llama3}, StreamSynth emphasizes experience accumulation across a task stream, without assuming a fixed or carefully designed task order. 
Feedback and knowledge acquired from historical tasks serve as informative inductive signals for future synthesis, enabling the model to transfer synthesis behavior to later and potentially different tasks. 
This setting better reflects realistic scenarios in which tasks evolve over time, task arrival order may be unpredictable, and effective synthesis benefits from leveraging prior experience rather than restarting from scratch \citep{Hallucination,Automatic_Instruction,Self-Instruct}.


Despite its potential, the StreamSynth setting introduces several fundamental challenges. 
First, most existing synthesis pipelines follow a generate-then-filter paradigm, where data construction is decoupled from explicit feedback signals. 
Without direct reward guidance, models cannot perceive why certain generations are preferable, limiting their ability to refine synthesis behavior. 
Second, beyond individual tasks, current approaches lack mechanisms to accumulate and organize synthesis experience across a task stream, making it difficult to develop stable, transferable, and generalizable synthesis capability over time, especially when later tasks differ from earlier ones. 
Finally, prior work predominantly evaluates synthesis at the level of individual samples, largely overlooking set-level diversity, which is crucial for constructing informative and reusable synthetic corpora in streaming settings.


To address these challenges, we propose \textbf{SynLearner}, a unified framework that enables models to \emph{learn to synthesize} under the StreamSynth setting (Figure~\ref{fig:framework}). 
SynLearner is designed to support synthesis learning over a task stream by encouraging experience reuse, stabilizing synthesis behavior across tasks, and explicitly balancing quality and diversity. 
It contains two key designs: \textbf{Diversity-Aware Initialization (DAI)}, which promotes broad exploration and diverse synthesis patterns at the beginning of each task; and \textbf{Hierarchical Reward Optimization (HRO)}, which provides explicit feedback at both the sample and set levels to guide synthesis behavior. 
Together, these designs form a \textbf{stream-oriented, feedback-driven learning loop} that enables synthesis strategies to be refined, reused, and transferred across tasks.


Our main contributions are as follows:
\begin{itemize}
    \item We formalize \textbf{StreamSynth}, a new setting that views synthetic data generation as an experience-driven process over task streams, aiming to study how synthesis experience can be accumulated, transferred, and generalized to future tasks.
    \item We propose \textbf{SynLearner}, a feedback-driven framework that enables synthesis models to \emph{learn to synthesize} from historical tasks, using reward-based optimization to refine generation behavior across evolving task streams.
    \item Extensive experiments show that SynLearner improves transferable synthesis capability across tasks, orders, model scales, and additional reasoning domains, demonstrating that synthesis experience learned from historical tasks can generalize beyond isolated data generation.
\end{itemize}

\section{Related Work}
\label{related_work}
\paragraph{LLM-based synthetic data generation.}
LLMs have been widely used for synthetic data generation in instruction tuning, reasoning, alignment, and data-efficient training~\citep{zihao_xue_tp,ours_1,skill_net}.
Representative studies include seed-based instruction bootstrapping~\citep{Self-Instruct}, instruction evolution~\citep{WizardLM}, and compact synthetic corpora for data-efficient learning~\citep{TinyStories}.
More specialized methods target mathematical reasoning and alignment, including concept-based problem synthesis~\citep{MathSmith}, difficulty-aware query generation~\citep{DART-Math}, multi-perspective mathematical rewriting~\citep{synthetic_metamath}, self-synthetic alignment~\citep{synthetic_1}, and rephrasing-based augmentation for language or multimodal reasoning~\citep{synthetic_rephrase,synthetic_rephrase1}.
While effective, these methods mainly treat synthesis as a static, task-specific process, without explicitly modeling how synthesis experience can be accumulated and reused across evolving tasks.

\paragraph{Continual and meta learning.}
Continual learning studies how models preserve predictive performance under sequential distribution shifts~\citep{continual_learning_method,continual_learning_method1,continual_learning_survery1,continual_learning_survery2}.
Recent LLM-oriented methods, such as GORP, FAPM, InsCL, and SEEKR, further explore parameter adaptation, replay, or knowledge retention across task streams~\citep{GORP,FAPM,replay-inscl,replay-seekr}.
Meta-learning instead aims to improve fast adaptation to new tasks by learning from task distributions or meta-training episodes~\citep{meta_learning_method1,meta_learning_method2,meta_learning_method3,meta_learning_method4}.
These paradigms are closely related to StreamSynth in their use of prior task experience, but their primary goals are predictive retention or fast adaptation rather than learning transferable data synthesis behavior.
\paragraph{Comparison with existing paradigms.}
As summarized in Table~\ref{tab:paradigm_comparison}, existing paradigms only partially satisfy the requirements of streaming synthesis.
Static synthesis focuses on task-specific data generation, while multi-task synthesis can exploit shared information across multiple observed tasks but usually assumes joint task access rather than sequential arrival.
Continual learning models sequential access, but mainly targets predictive retention rather than transferable synthesis behavior; meta-learning targets future adaptation but usually assumes access to task distributions or meta-training episodes.
In contrast, StreamSynth formulates synthesis as a feedback-driven process where historical experience is accumulated to acquire reusable synthesis behavior for future tasks.

\begin{table}[t]
\centering
\footnotesize
\setlength{\tabcolsep}{2.5pt}
\renewcommand{\arraystretch}{0.95}
\resizebox{\linewidth}{!}{
\begin{tabular}{lccccc}
\toprule
\textbf{Paradigm} 
& \textbf{Seq.} 
& \textbf{Synth.} 
& \textbf{Hist.} 
& \textbf{Feedback} 
& \textbf{Future} \\
& \textbf{Tasks} 
& \textbf{Obj.} 
& \textbf{Exp.} 
& \textbf{Opt.} 
& \textbf{Gen.} \\
\midrule
Static synthesis 
& \xmark & \cmark & \xmark & \xmark & \xmark \\

Multi-task synthesis
& \xmark & \cmark & \cmark & \xmark & \xmark \\

Continual learning 
& \cmark & \xmark & \cmark & \xmark & \xmark \\

Meta-learning 
& \xmark & \xmark & \cmark & \xmark & \cmark \\

\textbf{StreamSynth (Ours)} 
& \cmark & \cmark & \cmark & \cmark & \cmark \\
\bottomrule
\end{tabular}}
\caption{
Comparison with related paradigms.
``Seq.'', ``Synth.'', ``Hist.'', ``Feedback'', and ``Future'' denote sequential task modeling, synthesis-oriented objective, historical experience, feedback optimization, and future-task generalization.
}
\label{tab:paradigm_comparison}
\end{table}

\section{Problem Definition}
\label{sec:problem_definition}

We formalize synthetic data generation as a sequential synthesis learning problem, termed \textbf{StreamSynth}, where synthesis tasks arrive over time and the model adapts its generation behavior based on past experience.

\subsection{Formal Setting}
Let $\mathcal{T} = \{T_1, T_2, \dots, T_N\}$ denote a sequence of synthesis tasks.
At step $t$, task $T_t$ is defined by a prompt distribution $\mathcal{P}_t$ and an output space $\mathcal{Y}_t$:
\begin{equation}
\begin{aligned}
T_t &= (\mathcal{P}_t,\mathcal{Y}_t),\quad
p^{(t)}\sim\mathcal{P}_t,\\
x^{(t)} &\sim \mathcal{M}(p^{(t)}),\ x^{(t)}\in\mathcal{Y}_t .
\end{aligned}
\end{equation}
At task step $t$, the model has learned from previously observed tasks $\{T_1,\dots,T_{t-1}\}$ and receives the current task $T_t$, while future tasks remain unavailable.
The model must therefore update its synthesis behavior without access to future task information.

\subsection{Learning Objective}
The goal of StreamSynth is to optimize the synthesis model for future tasks in the stream.
We formalize this objective as maximizing the expected synthesis utility over unseen tasks:
\begin{equation}
\max_{\mathcal{M}_t} \ 
\mathbb{E}_{T_j \in \mathcal{T},\, j>t}
\big[ U(\mathcal{M}_t, T_j) \big],
\end{equation}
where $U(\mathcal{M}_t,T_j)$ evaluates the utility of synthetic data generated by model state $\mathcal{M}_t$ on a future task $T_j$, jointly considering sample-level quality and set-level diversity.

\begin{figure*}[t]
  \begin{center}
    \includegraphics[width=0.9\textwidth]{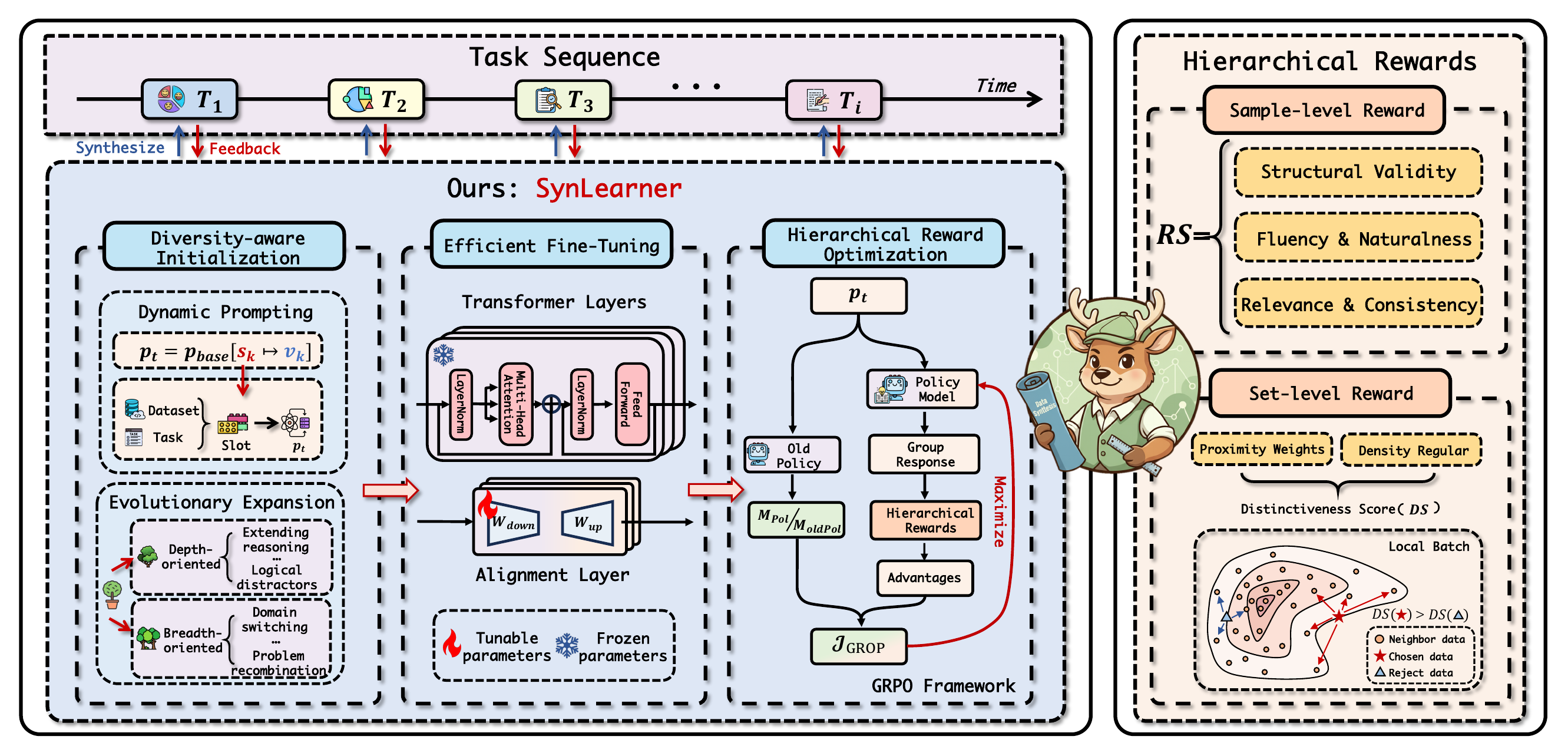}
  \end{center}
  \caption{
Overview of \textbf{SynLearner}.
Given a sequence of synthesis tasks, SynLearner initializes diverse synthesis conditions via dynamic prompting and evolutionary expansion, then optimizes the model through EFT and HRO with sample-level and set-level rewards.
This design enables the synthesis model to progressively improve generation quality and transfer its synthesis ability across tasks.
}
  \label{fig:framework}
\end{figure*}

\section{SynLearner Framework}
\label{method}

We introduce \emph{SynLearner} (Figure~\ref{fig:framework}), a framework for synthetic data generation under the StreamSynth setting.
It contains three functional components: Diversity-Aware Initialization (DAI), Efficient Fine-Tuning (EFT), and Hierarchical Reward Optimization (HRO).
DAI provides diverse task-specific starting conditions, while EFT and HRO form a unified optimization pipeline driven by dual-level reward signals.

\subsection{Diversity-Aware Initialization}
\label{sec:diversity_init}

In StreamSynth, the model must adapt to new tasks while preserving transferable synthesis strategies.
SynLearner therefore adopts \textbf{diversity-aware initialization} to expose the model to diverse and coherent synthesis conditions at each task stage.

We first use \textbf{dynamic prompting}, where a shared base template is instantiated with task-dependent slot values.
Let $p_{\text{base}}$ be the global template, $\mathcal{S}=\{s_1,\dots,s_K\}$ be controllable slots, and $\mathbf{v}_t=(v_{t,1},\dots,v_{t,K})$ be the slot assignment for task $T_t$.
The prompt is constructed as
\begin{equation}
p_t = p_{\text{base}}(\mathbf{v}_t),
\quad
v_{t,k} \sim \mathcal{V}_k,\ \forall k,
\quad \text{s.t. } \mathcal{C}_t .
\end{equation}
Here, $\mathcal{C}_t$ filters incoherent slot combinations, ensuring semantic consistency while preserving controllable diversity.

We further apply \textbf{evolutionary expansion} along two dimensions: depth-oriented evolution, which increases internal prompt complexity, and breadth-oriented evolution, which expands coverage across domains, styles, or problem forms.
The resulting prompt pool is aligned with the reward design in Section~\ref{sec:scoring}, linking diversity control with downstream feedback optimization.

\begin{table*}[ht!]
\centering
\footnotesize
\renewcommand{\arraystretch}{0.8} 
\resizebox{\textwidth}{!}{
\begin{tabular}{llcccccccc}
\toprule
\multirow{2}{*}{\textbf{Model}} & \multirow{2}{*}{\textbf{Methods}} 
& \multicolumn{2}{c}{\fontsize{8pt}{9pt}\selectfont \textbf{{Step 1:}} Yelp}  
& \multicolumn{2}{c}{\fontsize{8pt}{9pt}\selectfont \textbf{{Step 2:}} Amazon}
& \multicolumn{2}{c}{\fontsize{8pt}{9pt}\selectfont \textbf{{Step 3:}} Yahoo}
& \multicolumn{2}{c}{\fontsize{8pt}{9pt}\selectfont \textbf{{Step 4:}} MNLI}
\\
\cmidrule(lr){3-4} \cmidrule(lr){5-6} \cmidrule(lr){7-8} \cmidrule(lr){9-10} 
&
& \multicolumn{1}{c}{\fontsize{5pt}{6pt}\selectfont Acc (\%)}   
& \multicolumn{1}{c}{\fontsize{5pt}{6pt}\selectfont F1-Scores (\%)}  
& \multicolumn{1}{c}{\fontsize{5pt}{6pt}\selectfont Acc (\%)}   
& \multicolumn{1}{c}{\fontsize{5pt}{6pt}\selectfont F1-Scores (\%)}  
& \multicolumn{1}{c}{\fontsize{5pt}{6pt}\selectfont Acc (\%)}   
& \multicolumn{1}{c}{\fontsize{5pt}{6pt}\selectfont F1-Scores (\%)}   
& \multicolumn{1}{c}{\fontsize{5pt}{6pt}\selectfont Acc (\%)}   
& \multicolumn{1}{c}{\fontsize{5pt}{6pt}\selectfont F1-Scores (\%)}  
\\

\midrule
\multirow{12}{*}{\scriptsize LLaMA3.1-8B} 
& \multicolumn{9}{c}{\scriptsize No Training (Direct Synthesis)} \\ 
& \scriptsize Ori & \scriptsize 55.42  & \scriptsize 54.77 & \scriptsize 54.51  & \scriptsize 54.23  & \scriptsize 65.57  & \scriptsize 65.65 & \scriptsize 73.18  & \scriptsize 73.87  \\
& \scriptsize LLaMA3.1-8B & \scriptsize 59.87  & \scriptsize 60.00 & \scriptsize 57.46  & \scriptsize 57.76  & \scriptsize 64.87  & \scriptsize 65.37 & \scriptsize 76.92  & \scriptsize 78.62 \\
& \scriptsize Qwen2.5-7B & \scriptsize 62.08 & \scriptsize 62.17 & \scriptsize 57.86 & \scriptsize 58.03 & \scriptsize 64.16  & \scriptsize 64.05 & \scriptsize 78.52  & \scriptsize 77.66  \\
& \scriptsize LLaMA3.1-70B & \scriptsize 61.67 & \scriptsize 61.92 & \scriptsize 58.17 & \scriptsize 58.29  & \scriptsize 68.24  & \scriptsize 68.32 & \scriptsize 78.67  & \scriptsize 78.80  \\
& \scriptsize Deepseek & \scriptsize 65.16  & \scriptsize 65.43 & \scriptsize \textbf{60.22}  & \scriptsize \textbf{60.55}  & \scriptsize \underline{68.83}  & \scriptsize \underline{68.39} & \scriptsize 80.76  & \scriptsize \underline{80.66}  \\

\cmidrule(lr){2-10} 
& \multicolumn{9}{c}{\scriptsize With Training (Continual Learning-Based Synthesis)} \\ 
& \scriptsize GORP & \scriptsize 64.92  & \scriptsize65.23 & \scriptsize 58.74 & \scriptsize 59.12  & \scriptsize 67.83  & \scriptsize 67.92   & \scriptsize \underline{80.91} & \scriptsize 80.49 \\
& \scriptsize FAPM & \scriptsize \underline{65.89}  &  \scriptsize 65.33 & \scriptsize 58.42  & \scriptsize 58.64  & \scriptsize 66.45  & \scriptsize 66.60 & \scriptsize 78.48 & \scriptsize 78.80 \\
& \scriptsize InsCL & \scriptsize 65.46  & \scriptsize 65.42 & \scriptsize 58.39  & \scriptsize 58.73  & \scriptsize 66.38  & \scriptsize 66.84 & \scriptsize 79.53 & \scriptsize 79.10 \\
& \scriptsize SEEKR & \scriptsize 65.39  & \scriptsize \underline{65.65} & \scriptsize 58.36  & \scriptsize 58.71  & \scriptsize 66.59  & \scriptsize 66.54 & \scriptsize 80.06 & \scriptsize 79.99 \\
\cmidrule(lr){2-10} 
& \cellcolor{purple!20}\scriptsize Ours & \cellcolor{purple!20}\scriptsize \textbf{66.11} & \cellcolor{purple!20}\scriptsize \textbf{65.76} & \cellcolor{purple!20}\scriptsize \underline{60.09}  & \cellcolor{purple!20}\scriptsize \underline{60.39}  & \cellcolor{purple!20}\scriptsize \textbf{70.34}  & \cellcolor{purple!20}\scriptsize \textbf{70.11} & \cellcolor{purple!20}\scriptsize \textbf{82.50}  & \cellcolor{purple!20}\scriptsize \textbf{82.28}  \\

\midrule

\multirow{12}{*}{\scriptsize Qwen2.5-7B}
& \multicolumn{9}{c}{\scriptsize No Training (Direct Synthesis)} \\ 
&\scriptsize Ori & \scriptsize 58.08  & \scriptsize 54.48 & \scriptsize 55.41  & \scriptsize 55.47 & \scriptsize 65.03  & \scriptsize 65.09 & \scriptsize 85.50 & \scriptsize 85.37  \\
& \scriptsize LLaMA3.1-8B & \scriptsize 58.66 & \scriptsize 59.04  & \scriptsize 54.49  & \scriptsize 54.70  & \scriptsize 66.88  & \scriptsize 66.93 & \scriptsize 85.98 & \scriptsize 85.96 \\
& \scriptsize Qwen2.5-7B & \scriptsize 62.01 & \scriptsize 62.41 & \scriptsize 56.39 & \scriptsize 56.54 & \scriptsize 66.20  & \scriptsize 65.79 & \scriptsize 86.00  & \scriptsize 85.94 \\
& \scriptsize LLaMA3.1-70B & \scriptsize 60.74 & \scriptsize 61.06  & \scriptsize 56.61 & \scriptsize 56.88  & \scriptsize 67.17  & \scriptsize 67.19  & \scriptsize \underline{86.01}   & \scriptsize \underline{85.96} \\
& \scriptsize Deepseek & \scriptsize \underline{64.86} & \scriptsize \underline{65.12} & \scriptsize 57.29 & \scriptsize 57.49 & \scriptsize \underline{69.24}  & \scriptsize \textbf{69.05} & \scriptsize 85.81  & \scriptsize 85.67 \\

\cmidrule(lr){2-10} 
& \multicolumn{9}{c}{\scriptsize With Training (Continual Learning-Based Synthesis)} \\ 

& \scriptsize GORP & \scriptsize \textbf{64.87}  & \scriptsize \textbf{64.98} & \scriptsize 56.82  & \scriptsize 57.09  & \scriptsize 68.25  & \scriptsize 68.42 & \scriptsize 85.53 & \scriptsize 85.28 \\
& \scriptsize FAPM & \scriptsize 63.88  & \scriptsize 64.16 & \scriptsize \underline{57.64}  & \scriptsize \underline{57.91}  & \scriptsize 67.71  & \scriptsize 67.82 & \scriptsize 84.88 & \scriptsize 85.02 \\
& \scriptsize InsCL & \scriptsize 61.53  & \scriptsize 60.29 & \scriptsize 57.54  & \scriptsize 57.80  & \scriptsize 67.53  & \scriptsize 67.70 & \scriptsize 85.79 & \scriptsize 85.79 \\
& \scriptsize SEEKR & \scriptsize 63.11  & \scriptsize 63.30 & \scriptsize 57.18  & \scriptsize 57.39  & \scriptsize 67.26  & \scriptsize 67.35 & \scriptsize 85.67 & \scriptsize 85.44 \\
\cmidrule(lr){2-10} 
& \cellcolor{purple!20}\scriptsize Ours & \cellcolor{purple!20}\scriptsize 63.91  & \cellcolor{purple!20}\scriptsize 64.13 & \cellcolor{purple!20}\scriptsize \textbf{58.59}  & \cellcolor{purple!20}\scriptsize \textbf{58.66}  & \cellcolor{purple!20}\scriptsize  \textbf{69.25} & \cellcolor{purple!20}\scriptsize \underline{68.94} & \cellcolor{purple!20}\scriptsize \textbf{86.56}   & \cellcolor{purple!20}\scriptsize \textbf{86.47}  \\

\bottomrule
\end{tabular}}
\caption{
Comparison of synthesis performance across streaming tasks using the \textbf{StreamSynth} pipeline. 
Results show \textbf{Accuracy (Acc)} and \textbf{F1-scores} for each task in the stream 
(\textit{Yelp} $\rightarrow$ \textit{Amazon} $\rightarrow$ \textit{Yahoo} $\rightarrow$ \textit{MNLI}) 
under two backbone models (LLaMA3.1-8B and Qwen2.5-7B). 
Both direct (no-training) and continual learning-based synthesis approaches are evaluated under identical streaming settings. 
Bold numbers indicate the best performance within each model group, and underlined numbers indicate the second-best performance.
}
\label{tab:main}
\end{table*}

\subsection{Dual-level Scoring Strategy}
\label{sec:scoring}

To support feedback-driven synthesis learning, we define dual-level rewards that jointly evaluate sample quality and set-level diversity.

\paragraph{Sample-level Scoring.}
For a synthesized sample $x^{(t)}$ under task $T_t$, the sample reward is
\begin{equation}
\begin{aligned}
RS_t(x^{(t)}) =
&\gamma_1 S_{\text{struct}}(x^{(t)})
+\gamma_2 S_{\text{fluent}}(x^{(t)}) \\
&+\gamma_3 S^{(t)}_{\text{rel}}(x^{(t)}),
\end{aligned}
\end{equation}
where $\gamma_1,\gamma_2,\gamma_3\ge 0$, $\gamma_1+\gamma_2+\gamma_3=1$, and all scores are normalized to $[0,1]$.
Here, $S_{\text{struct}}$ checks task-agnostic format validity, $S_{\text{fluent}}$ measures linguistic naturalness, and $S^{(t)}_{\text{rel}}$ evaluates task-specific semantic alignment.
Concretely,
\begin{equation}
\begin{aligned}
S_{\text{fluent}}(x^{(t)}) &=
\alpha \mathrm{LM}(x^{(t)})\\
&\quad+(1-\alpha)\mathrm{style}(x^{(t)}), \\
S^{(t)}_{\text{rel}}(x^{(t)}) &=
\theta p_{\tau_t}(\ell^{(t)}|x^{(t)}) \\
&\quad +(1-\theta)m_t(x^{(t)},\ell^{(t)}).
\end{aligned}
\end{equation}
where $\ell^{(t)}$ denotes the target label, and $\alpha,\theta\in[0,1]$ control the corresponding trade-offs.

\paragraph{Set-level Scoring.}
To encourage diversity, we compute a local distinctiveness score within a mini-batch 
$\mathcal{B}_t=\{x^{(t)}_1,\dots,x^{(t)}_m\}$.
For each sample $x^{(t)}_i\in\mathcal{B}_t$, we compare it with other samples in the same mini-batch:
\begin{equation}
\begin{aligned}
\mathrm{sim}_{ij}^{(t)} &=
\cos(\mathcal{E}(x^{(t)}_i),\mathcal{E}(x^{(t)}_j)), \quad j\ne i, \\
w_{ij}^{(t)} &=
\frac{\exp(\mathrm{sim}_{ij}^{(t)}/\tau)}
{\sum_{\ell\ne i}\exp(\mathrm{sim}_{i\ell}^{(t)}/\tau)}, \\
D_t(x^{(t)}_i) &= \sum_{j\ne i} w_{ij}^{(t)} \mathrm{sim}_{ij}^{(t)}, \\
DS_t(x^{(t)}_i) &= \exp(-kD_t(x^{(t)}_i)),
\end{aligned}
\end{equation}
where $\tau>0$ is a temperature parameter and $k>0$ controls the diversity penalty.
Higher local density leads to a lower distinctiveness score, discouraging redundant synthetic samples.

\subsection{Hierarchical Training Procedure}
\label{sec:training}

SynLearner optimizes synthesis behavior through EFT followed by HRO.

\paragraph{Stage 1: Efficient Fine-Tuning.}
For each incoming task $T_t$, we construct $\mathcal{D}^{(t)}_{\text{sft}}$ using dynamically prompted samples and optional real examples.
Standard supervised fine-tuning then produces an EFT-initialized model $\widetilde{\mathcal{M}}_t$ from the previous model $\mathcal{M}_{t-1}$.
This stage provides basic task-following and structurally valid generation ability before feedback-based optimization.

\paragraph{Stage 2: Hierarchical Reward Optimization.}
Starting from $\widetilde{\mathcal{M}}_t$, SynLearner performs reinforcement learning using hierarchical rewards.
Given a prompt $p^{(t)}\sim\mathcal{P}_t$, the model generates
$x^{(t)}\sim\widetilde{\mathcal{M}}_t(p^{(t)})$.
We then compute
\begin{equation}
r^{(t)}_{\text{sample}}=RS_t(x^{(t)}),
\quad
r^{(t)}_{\text{set}}=DS_t(x^{(t)}),
\end{equation}
\begin{equation}
r^{(t)}_{\text{total}}
=\lambda r^{(t)}_{\text{sample}}
+(1-\lambda)r^{(t)}_{\text{set}},
\end{equation}
where $\lambda\in[0,1]$ balances sample-level quality and set-level diversity.
The model is updated as
\begin{equation}
\mathcal{M}_t \leftarrow
\mathrm{RL\text{-}Update}(\widetilde{\mathcal{M}}_t,p^{(t)},x^{(t)},r^{(t)}_{\text{total}}).
\end{equation}
During optimization, we compute set-level scores within mini-batches and incrementally update the candidate pool, making the reward signal stable and informative.
As tasks arrive sequentially, the model sequence $\{\mathcal{M}_t\}$ accumulates synthesis experience for future tasks.

\section{Experiments and Analysis}
\label{experiments}

\subsection{Experimental Setup}
\label{sec:exp_setup}
\paragraph{Benchmarks.}
We evaluate \textbf{SynLearner} under the \textbf{StreamSynth} setting on four real-world NLP benchmarks: \textbf{Yelp}, \textbf{Yahoo}, and \textbf{Amazon} from the StandardCL benchmark~\citep{stand_cl_3_dataset}, as well as \textbf{MNLI} from GLUE~\citep{glue-mnli}.
These datasets cover diverse language understanding scenarios, including review understanding, topic-oriented text understanding, and natural language inference.
All experiments are conducted using instruction-tuned LLMs, including LLaMA3.1-8B~\citep{llama3} and Qwen2.5-7B~\citep{qwen2_5}, which are used consistently for both data synthesis and downstream training.


\begin{table*}[ht!]
\centering
\footnotesize
\renewcommand{\arraystretch}{0.8} 
\resizebox{\textwidth}{!}{
\begin{tabular}{llcccccccc}
\toprule
\multirow{2}{*}{\textbf{Model}} & \multirow{2}{*}{\textbf{Methods}} 
& \multicolumn{2}{c}{\fontsize{8pt}{9pt}\selectfont \textbf{{Step 1:}} Yelp}  
& \multicolumn{2}{c}{\fontsize{8pt}{9pt}\selectfont \textbf{{Step 2:}} Amazon}
& \multicolumn{2}{c}{\fontsize{8pt}{9pt}\selectfont \textbf{{Step 3:}} Yahoo}
& \multicolumn{2}{c}{\fontsize{8pt}{9pt}\selectfont \textbf{{Step 4:}} MNLI}
\\
\cmidrule(lr){3-4} \cmidrule(lr){5-6} \cmidrule(lr){7-8} \cmidrule(lr){9-10} 
&
& \multicolumn{1}{c}{\fontsize{5pt}{6pt}\selectfont Acc (\%)}   
& \multicolumn{1}{c}{\fontsize{5pt}{6pt}\selectfont F1-Scores (\%)}  
& \multicolumn{1}{c}{\fontsize{5pt}{6pt}\selectfont Acc (\%)}   
& \multicolumn{1}{c}{\fontsize{5pt}{6pt}\selectfont F1-Scores (\%)}  
& \multicolumn{1}{c}{\fontsize{5pt}{6pt}\selectfont Acc (\%)}   
& \multicolumn{1}{c}{\fontsize{5pt}{6pt}\selectfont F1-Scores (\%)}   
& \multicolumn{1}{c}{\fontsize{5pt}{6pt}\selectfont Acc (\%)}   
& \multicolumn{1}{c}{\fontsize{5pt}{6pt}\selectfont F1-Scores (\%)}  
\\
\midrule
\multirow{11}{*}{\scriptsize LLaMA3.1-8B}  
& \multicolumn{9}{c}{\scriptsize Diversity Prompt Ablation} \\
& \scriptsize Ours w/o Dynamic Prompt & \scriptsize 60.59 & \scriptsize 60.54 & \scriptsize 56.78 & \scriptsize 56.68 & \scriptsize 66.53 & \scriptsize 66.55 & \scriptsize 79.66 & \scriptsize 79.86 \\
& \scriptsize Ours w/o Evolutionary Expansion & \scriptsize 63.16 & \scriptsize 63.43 & \scriptsize 57.83 & \scriptsize 58.05 & \scriptsize 66.99 & \scriptsize 67.24 & \scriptsize 78.31 & \scriptsize 78.34 \\
\cmidrule(lr){2-10}
& \multicolumn{9}{c}{\scriptsize RL Component Ablation} \\
& \scriptsize Ours w/o Structural Reward & \scriptsize 64.54 & \scriptsize 64.79 & \scriptsize 59.07 & \scriptsize 59.33 & \scriptsize 69.54 & \scriptsize 69.59 & \scriptsize 78.59 & \scriptsize 78.54 \\
& \scriptsize Ours w/o Fluency Reward & \scriptsize 65.05 & \scriptsize 65.32 & \scriptsize 58.70 & \scriptsize 58.96 & \scriptsize 69.95 & \scriptsize 69.81 & \scriptsize 79.60 & \scriptsize 79.76 \\
& \scriptsize Ours w/o Relevance Reward & \scriptsize 64.21 & \scriptsize 64.48 & 
\scriptsize 59.07 & \scriptsize 59.31 & \scriptsize 69.68 & \scriptsize 69.74 & \scriptsize 78.22 & \scriptsize 78.36 \\
& \scriptsize Ours w/o Set-level Reward & \scriptsize 64.22 & \scriptsize 64.48 & 
\scriptsize 59.20 & \scriptsize 59.49 & \scriptsize 70.08 & \scriptsize 69.88 & \scriptsize 79.41 & \scriptsize 79.56 \\
& \scriptsize Ours w/o GRPO & \scriptsize 62.58 & \scriptsize 62.92 & \scriptsize 58.53 & \scriptsize 58.84 & \scriptsize 67.38 & \scriptsize 67.45 & \scriptsize 78.29 & \scriptsize 78.22 \\

\cmidrule(lr){2-10}
& \cellcolor{purple!20}\scriptsize Ours & \cellcolor{purple!20}\scriptsize \textbf{66.11} & \cellcolor{purple!20}\scriptsize \textbf{65.76} & \cellcolor{purple!20}\scriptsize \textbf{60.09}  & \cellcolor{purple!20}\scriptsize \textbf{60.39}  & \cellcolor{purple!20}\scriptsize \textbf{70.34}  & \cellcolor{purple!20}\scriptsize \textbf{70.11} & \cellcolor{purple!20}\scriptsize \textbf{82.50}  & \cellcolor{purple!20}\scriptsize \textbf{82.28}  \\
\bottomrule
\end{tabular}}
\caption{Ablation of Core Components in SynLearner Under the StreamSynth Setting.
We ablate Diversity-Aware Prompting and Hierarchical Reward Optimization to assess their roles in synthesis learning.}
\label{tab:ablation_wo_prompt_rl}
\end{table*}

\begin{table*}[ht!]
\centering
\footnotesize
\renewcommand{\arraystretch}{0.8} 
\resizebox{\textwidth}{!}{
\begin{tabular}{llcccccccccc}
\toprule
\multirow{2}{*}{\textbf{Model}} & \multirow{2}{*}{\textbf{Methods}} 
& \multicolumn{3}{c}{\fontsize{8pt}{9pt}\selectfont \textbf{{Step 1$\rightarrow$2:}} Amazon }  
& \multicolumn{3}{c}{\fontsize{8pt}{9pt}\selectfont \textbf{{Step 2$\rightarrow$3:}} Yahoo }
& \multicolumn{3}{c}{\fontsize{8pt}{9pt}\selectfont \textbf{{Step 3$\rightarrow$4:}} MNLI }
\\
\cmidrule(lr){3-5} \cmidrule(lr){6-8} \cmidrule(lr){9-11} 
&
& \multicolumn{1}{c}{\fontsize{5pt}{6pt}\selectfont Acc (\%)}   
& \multicolumn{1}{c}{\fontsize{5pt}{6pt}\selectfont F1-Scores (\%)}  
& \multicolumn{1}{c}{\fontsize{5pt}{6pt}\selectfont $\Delta$(\%)}
& \multicolumn{1}{c}{\fontsize{5pt}{6pt}\selectfont Acc (\%)}   
& \multicolumn{1}{c}{\fontsize{5pt}{6pt}\selectfont F1-Scores (\%)}  
& \multicolumn{1}{c}{\fontsize{5pt}{6pt}\selectfont $\Delta$(\%)}
& \multicolumn{1}{c}{\fontsize{5pt}{6pt}\selectfont Acc (\%)}   
& \multicolumn{1}{c}{\fontsize{5pt}{6pt}\selectfont F1-Scores (\%)}  
& \multicolumn{1}{c}{\fontsize{5pt}{6pt}\selectfont $\Delta$(\%)}
\\
\midrule
\multirow{5}{*}{\scriptsize LLaMA3.1-8B} 
& \scriptsize GORP    & \scriptsize 56.74 & \scriptsize 55.97 & \scriptsize -0.72 & \scriptsize 64.29 & \scriptsize 64.16 & \scriptsize -0.58 & \scriptsize 75.71 & \scriptsize 76.03 & \scriptsize -1.21 \\
& \scriptsize FAPM    & \scriptsize 57.61  & \scriptsize 57.61 & \scriptsize +0.15 & \scriptsize 65.36 & \scriptsize 65.41 & \scriptsize +0.49 & \scriptsize 77.17 & \scriptsize 77.27 & \scriptsize +0.25 \\
& \scriptsize InsCL   & \scriptsize 57.32 & \scriptsize 56.79 & \scriptsize -0.14 & \scriptsize 65.21 & \scriptsize 64.81 & \scriptsize +0.34 & \scriptsize 76.21 & \scriptsize 76.54 & \scriptsize -0.71 \\
& \scriptsize SEEKR   & \scriptsize 58.01 & \scriptsize 58.13 & \scriptsize +0.55 & \scriptsize 64.08 & \scriptsize 64.02 & \scriptsize -0.79 & \scriptsize 76.42 & \scriptsize 76.60 & \scriptsize -0.5 \\
& \cellcolor{purple!20}\scriptsize Ours & \cellcolor{purple!20}\scriptsize \textbf{58.76} &  \cellcolor{purple!20}\scriptsize \textbf{58.15} &  \cellcolor{purple!20}\scriptsize \textbf{+1.3} &  
\cellcolor{purple!20}\scriptsize \textbf{68.89} & 
\cellcolor{purple!20}\scriptsize \textbf{68.88} & 
\cellcolor{purple!20}\scriptsize \textbf{+4.02} & 
\cellcolor{purple!20}\scriptsize \textbf{79.72} & 
\cellcolor{purple!20}\scriptsize \textbf{79.53} & 
\cellcolor{purple!20}\scriptsize  \textbf{+2.8}  \\
\midrule
\multirow{5}{*}{\scriptsize Qwen2.5-7B}
& \scriptsize GORP    & \scriptsize 53.32  & \scriptsize 53.46 & \scriptsize -3.07 & \scriptsize 63.18 & \scriptsize 62.99 & \scriptsize -3.02 & \scriptsize 83.05 & \scriptsize 82.31 & \scriptsize -2.95 \\
& \scriptsize FAPM    & \scriptsize 55.55 & \scriptsize 55.77 & \scriptsize -0.84 & \scriptsize 65.93 & \scriptsize 65.47 & \scriptsize -0.27 & \scriptsize 82.72 & \scriptsize 82.17 & \scriptsize -3.28 \\
& \scriptsize InsCL   & \scriptsize 54.84 & \scriptsize 55.13 & \scriptsize -1.55 & \scriptsize 65.36 & \scriptsize 65.06 & \scriptsize -0.84 & \scriptsize 82.73 & \scriptsize 82.41 & \scriptsize -3.27 \\
& \scriptsize SEEKR   & \scriptsize 55.93 & \scriptsize 56.23 & \scriptsize -0.46 & \scriptsize 65.47 & \scriptsize 65.29 & \scriptsize -0.73 & \scriptsize 83.97 & \scriptsize 83.58 & \scriptsize -2.03 \\
& \cellcolor{purple!20}\scriptsize Ours    & \cellcolor{purple!20}\scriptsize \textbf{57.74} &  \cellcolor{purple!20}\scriptsize \textbf{57.92} &  \cellcolor{purple!20}\scriptsize \textbf{+1.35} & 
\cellcolor{purple!20}\scriptsize \textbf{67.79} & 
\cellcolor{purple!20}\scriptsize \textbf{67.30} & 
\cellcolor{purple!20}\scriptsize \textbf{+1.59} & 
\cellcolor{purple!20}\scriptsize \textbf{85.65} & 
\cellcolor{purple!20}\scriptsize \textbf{85.45} & 
\cellcolor{purple!20}\scriptsize \textbf{-0.35} 
\\
\bottomrule
\end{tabular}}
\caption{
Forward-transfer evaluation for cross-dataset generalization.
For each transition in the stream, the model learned from earlier tasks is directly evaluated on the next unseen target task without additional target-task training.
The $\Delta$(\%) column denotes the absolute accuracy difference from the corresponding direct synthesis baseline under the same backbone and target dataset.
}
\label{tab:next_prediction}
\end{table*}

\paragraph{Baselines and Evaluation.}
We compare \textbf{SynLearner} with three groups of representative baselines: 
(1) \textbf{direct synthesis} methods that generate synthetic data via prompt-only LLMs without parameter updates; 
(2) \textbf{parameter-efficient continual adaptation} methods, including GORP~\citep{GORP} and FAPM~\citep{FAPM}; and 
(3) \textbf{replay- and retention-based adaptation} methods, including InsCL~\citep{replay-inscl} and SEEKR~\citep{replay-seekr}. 
For all methods, synthetic samples are mixed with original training data at a fixed ratio, and downstream performance is evaluated on standard test splits.
Additional experimental details are provided in Appendix~\ref{appendix:exp_details}.

\begin{figure*}[t]
  \centering
  \includegraphics[width=1.0\linewidth]{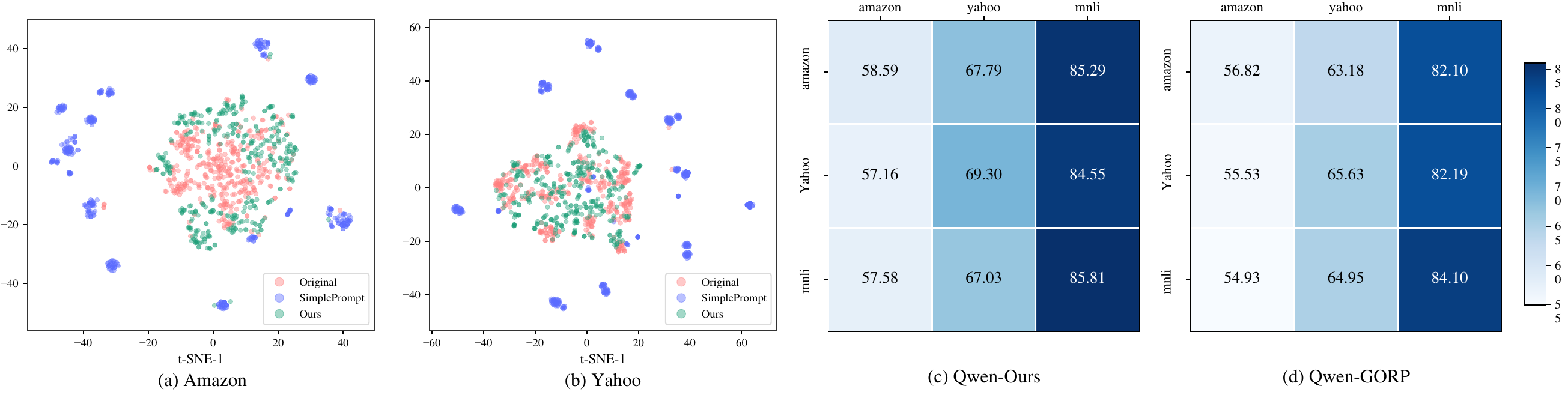}
  \caption{
  Selected qualitative analyses of synthesis diversity and cross-stage generalization.
  Subfigures (a) and (b) show t-SNE visualizations on Amazon and Yahoo, comparing original samples, simple-prompt synthesis, and SynLearner.
  Subfigures (c) and (d) show transfer heatmaps on Qwen, comparing SynLearner with GORP.
  In each heatmap, rows denote model states after sequentially learning up to the corresponding task in the stream, and columns denote target evaluation datasets.
  Darker cells indicate higher downstream accuracy.
  }
  \label{fig:qualitative_selected}
\end{figure*}

\subsection{Main Results Analysis} Table~\ref{tab:main} presents the main results of \textbf{SynLearner} under the \textbf{StreamSynth} setting across four sequential tasks (\textit{Yelp}, \textit{Amazon}, \textit{Yahoo}, and \textit{MNLI}) on LLaMA3.1-8B and Qwen2.5-7B. Overall, SynLearner achieves the strongest or highly competitive performance across both backbones and all task stages. Compared with direct synthesis and sequential adaptation baselines, SynLearner consistently improves downstream performance, showing that feedback-driven synthesis learning can better exploit historical synthesis experience instead of treating each task independently. This trend supports our central hypothesis that synthesis behavior can be accumulated and reused across a task stream, rather than being learned separately for each dataset. More specifically, SynLearner shows clearer advantages on later tasks, where transfer from previous synthesis experience becomes more important. On LLaMA3.1-8B, it achieves the best results on \textit{Yelp}, \textit{Yahoo}, and \textit{MNLI}, while remaining competitive on \textit{Amazon}. On Qwen2.5-7B, although the gains on early tasks are less pronounced, SynLearner becomes stronger as the stream progresses and achieves the best or near-best results on later tasks. The relatively smaller gains on some LLaMA3.1-8B intermediate datasets may be due to strong prompt-based baselines already producing high-quality synthetic data, leaving less room for additional improvement. This suggests that SynLearner is more beneficial when later tasks require transferable synthesis strategies beyond surface-level pattern generation.

\subsection{Ablation Study}
\label{sec:ablation}

We conduct comprehensive ablation studies under the \textbf{StreamSynth} setting to analyze each component’s contribution in \textbf{SynLearner} and better understand its learning behavior. Specifically, our ablations address three key questions:

\begin{questionbox}
\textbf{Q1:} Whether each core component of SynLearner is necessary for effective synthesis learning?

\textbf{Q2:} Whether SynLearner can acquire transferable and general synthesis capability under streaming tasks?

\textbf{Q3:} Whether the framework remains robust under different model scales?
\end{questionbox}

Unless otherwise specified, all experiments follow the same setup as the main results.





\paragraph{\textbf{Q1: Component Necessity — Prompt and Reward Design.}}
We first examine if SynLearner’s core components are necessary for effective synthesis learning.
As shown in Table~\ref{tab:ablation_wo_prompt_rl}, removing either diversity-aware prompting or reward-based optimization consistently degrades performance.
Prompt ablations weaken synthesis diversity, while reward ablations reduce quality, relevance, or set-level coverage.
Removing GRPO leads to the largest drop, confirming feedback-guided optimization is essential beyond supervised synthesis alone.

Figure~\ref{fig:qualitative_selected}(a,b) further provides selected t-SNE visualizations.
\textit{SimplePrompt} denotes a basic few-shot prompting strategy without dynamic or evolutionary prompt expansion.
Compared with SimplePrompt, SynLearner covers broader semantic regions while maintaining better alignment with the original data distribution, indicating improved diversity and distributional compatibility.
Full t-SNE results are provided in Figure~\ref{fig:tsne_distribution_full} of Appendix~\ref{app:extra_exp}.

\paragraph{\textbf{Q2: Learning Dynamics under StreamSynth.}}
We next examine if SynLearner gains transferable synthesis capability rather than fitting individual tasks.
Table~\ref{tab:next_prediction} shows that SynLearner consistently outperforms baselines when models trained on earlier tasks are directly evaluated on future tasks, demonstrating effective forward transfer.

Figure~\ref{fig:qualitative_selected}(c,d) provides selected heatmap analyses of cross-stage generalization.
In each heatmap, rows denote model states after sequentially learning up to the corresponding task, while columns denote target evaluation datasets.
Compared with GORP, SynLearner achieves stronger performance across both seen and unseen targets, suggesting that it learns synthesis behaviors that remain useful throughout the task stream.
Complete heatmap results are provided in Figure~\ref{fig:transfer_heatmap_full} of Appendix~\ref{app:extra_exp}.

\begin{figure*}[t!]
  \hspace*{\fill} 
  \includegraphics[width=1.00\linewidth]{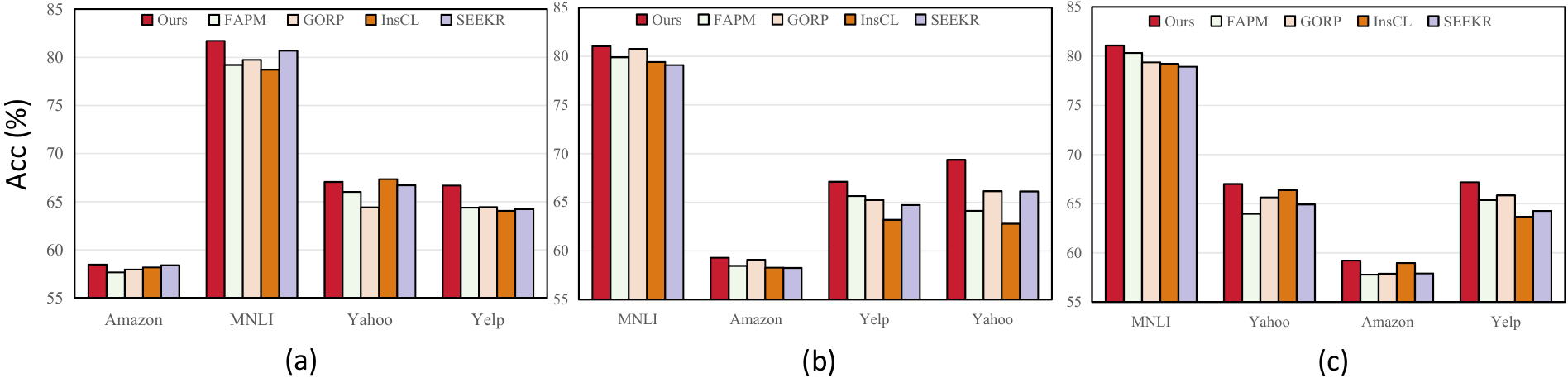}
  \hspace*{\fill} 
  \caption{Effect of task order on synthesis performance under different task stream permutations.
Subfigures (a), (b), and (c) correspond to three task orders:
Order1 (Amazon $\rightarrow$ MNLI $\rightarrow$ Yahoo $\rightarrow$ Yelp),
Order2 (MNLI $\rightarrow$ Amazon $\rightarrow$ Yelp $\rightarrow$ Yahoo),
and Order3 (MNLI $\rightarrow$ Yahoo $\rightarrow$ Amazon $\rightarrow$ Yelp).}
  \label{fig:task_order}
\end{figure*}

\begin{figure*}[t!]
  \hspace*{\fill} 
  \includegraphics[width=1.00\linewidth]{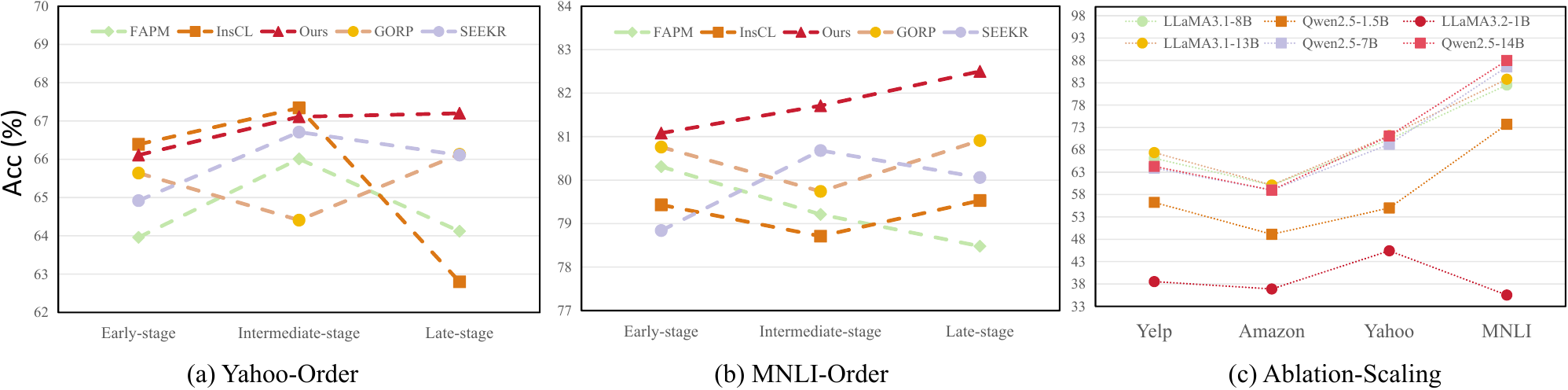}
  \hspace*{\fill} 
  \caption{Effect of task position in the stream on synthesis performance.
Subfigures (a) and (b) report results on two task streams, where tasks are grouped into \emph{Early}, \emph{Intermediate}, and \emph{Late} stages according to their positions in the stream.
Subfigure (c) analyzes the impact of model scale.}
  \label{fig:order_analyze}
\end{figure*}

\begin{table}[t]
\centering
\footnotesize
\renewcommand{\arraystretch}{0.9}
\resizebox{\linewidth}{!}{
\begin{tabular}{llcccc}
\toprule
\multirow{2}{*}{\textbf{Model}} 
& \multirow{2}{*}{\textbf{Method}} 
& \multicolumn{2}{c}{\textbf{GSM8K}} 
& \multicolumn{2}{c}{\textbf{MATH-500}} \\
\cmidrule(lr){3-4} \cmidrule(lr){5-6}
& & Acc (\%) & $\Delta$ (\%) & Acc (\%) & $\Delta$ (\%) \\
\midrule

\multirow{2}{*}{\scriptsize LLaMA3.1-8B}
& \scriptsize Ori  
& \scriptsize 70.20 & \scriptsize -- 
& \scriptsize 26.40 & \scriptsize -- \\

& \scriptsize Ours 
& \scriptsize \textbf{73.01} & \scriptsize +2.81
& \scriptsize \textbf{34.20} & \scriptsize +7.80 \\

\midrule

\multirow{2}{*}{\scriptsize Qwen2.5-7B}
& \scriptsize Ori  
& \scriptsize 74.75 & \scriptsize -- 
& \scriptsize 35.60 & \scriptsize -- \\

& \scriptsize Ours 
& \scriptsize \textbf{78.69} & \scriptsize +3.94
& \scriptsize \textbf{47.80} & \scriptsize +12.20 \\

\bottomrule
\end{tabular}}
\caption{
Step 5 generalization results on mathematical reasoning tasks.
GSM8K and MATH-500 are appended after the original StreamSynth sequence, and $\Delta$(\%) denotes the absolute accuracy gain over Ori.
}
\label{tab:math_generalization}
\end{table}

\textbf{Task Order and Stream Position Effects.}
We further evaluate SynLearner under three task stream permutations.
Figure~\ref{fig:task_order} shows that SynLearner remains consistently competitive across different orders, indicating robustness to task sequencing.
Figure~\ref{fig:order_analyze}(a,b) provides a finer-grained stage-level analysis by grouping tasks into \emph{Early}, \emph{Intermediate}, and \emph{Late} positions.
SynLearner shows clearer improvements when tasks appear later in the stream, suggesting that synthesis experience is accumulated and reused over time.

We also append GSM8K~\citep{gsm8k} and MATH-500~\citep{math500} as Step 5 reasoning tasks to examine cross-task-type generalization.
Table~\ref{tab:math_generalization} shows consistent gains over \textbf{Ori}, indicating that learned synthesis experience can transfer beyond the original task stream.
Because these reasoning benchmarks differ from the original language understanding tasks in structure and difficulty, the gains suggest that SynLearner learns transferable synthesis behavior rather than dataset-specific generation patterns.

\paragraph{\textbf{Q3: Robustness and Scalability.}}
Finally, we evaluate robustness with respect to model scale.
As shown in Figure~\ref{fig:order_analyze}(c), performance improves substantially from 1B--1.5B models to 7B--8B models, while further scaling to 13B--14B brings only limited gains.
This suggests that SynLearner benefits from stronger backbones, but its effectiveness is not solely driven by parameter scaling.

Overall, these ablations show that SynLearner benefits from well-aligned prompt design, feedback-based reward optimization, transferable synthesis learning under task streams, and robustness across model scales.

\section{Conclusion}
In this paper, we introduced \textbf{StreamSynth}, a new setting that studies synthetic data generation as an experience-driven process over task streams.
We further proposed \textbf{SynLearner}, a feedback-driven framework that enables synthesis models to learn reusable synthesis behavior from historical tasks.
Extensive experiments show that SynLearner improves synthesis quality, transferability, and generalization across diverse benchmarks.
These results demonstrate the feasibility of learning transferable synthesis capability through task-stream experience.
Future work will explore multimodal extensions and theoretical analysis of synthesis transfer.


\section*{Limitations}
While this work demonstrates the feasibility of synthetic data generation under streaming task settings, several limitations remain. 
First, although SynLearner accumulates experience across tasks, historical information is leveraged implicitly through reward-driven parameter updates, rather than being explicitly modeled over time. 
As a result, long-range temporal dependencies and fine-grained task evolution patterns may not be fully captured. 
Second, our evaluation mainly focuses on task streams with moderate semantic relatedness, aiming to verify whether synthesis experience from previous tasks can provide information gain for subsequent ones. 
The effectiveness of SynLearner under more heterogeneous or substantially different task domains is therefore not fully explored, which limits the current assessment of its generality.

\bibliography{anthology,custom}

\begin{thebibliography}{41}
\providecommand{\natexlab}[1]{#1}

\bibitem[{Bi et~al.(2026)Bi, Hu, Chen, Chen, Deng, Xue, Wang, Shen, Zhang, and Lou}]{ours_1}
Zhen Bi, Zhenlin Hu, Xueshu Chen, Mingyang Chen, Cheng Deng, Yida Xue, Zhen Wang, Qing Shen, Ningyu Zhang, and Jungang Lou. 2026.
\newblock \href {https://arxiv.org/abs/2509.24836} {Logical structure as knowledge: Enhancing llm reasoning via structured logical knowledge density estimation}.
\newblock \emph{Preprint}, arXiv:2509.24836.

\bibitem[{Choi et~al.(2025)Choi, Asif, Han, Willes, and Krishnan}]{meta_learning_method4}
Younwoo Choi, Muhammad~Adil Asif, Ziwen Han, John Willes, and Rahul Krishnan. 2025.
\newblock \href {https://openreview.net/forum?id=FS2nukC2jv} {Teaching {LLM}s how to learn with contextual fine-tuning}.
\newblock In \emph{The Thirteenth International Conference on Learning Representations}.

\bibitem[{Cobbe et~al.(2021)Cobbe, Kosaraju, Bavarian, Chen, Jun, Kaiser, Plappert, Tworek, Hilton, Nakano, Hesse, and Schulman}]{gsm8k}
Karl Cobbe, Vineet Kosaraju, Mohammad Bavarian, Mark Chen, Heewoo Jun, Lukasz Kaiser, Matthias Plappert, Jerry Tworek, Jacob Hilton, Reiichiro Nakano, Christopher Hesse, and John Schulman. 2021.
\newblock \href {https://arxiv.org/abs/2110.14168} {Training verifiers to solve math word problems}.
\newblock \emph{Preprint}, arXiv:2110.14168.

\bibitem[{DeepSeek-AI(2025)}]{Deepseek}
DeepSeek-AI. 2025.
\newblock \href {https://arxiv.org/abs/2501.12948} {Deepseek-r1: Incentivizing reasoning capability in llms via reinforcement learning}.
\newblock \emph{Preprint}, arXiv:2501.12948.

\bibitem[{Deng et~al.(2023)Deng, Zhang, Chen, and Gu}]{synthetic_rephrase}
Yihe Deng, Weitong Zhang, Zixiang Chen, and Quanquan Gu. 2023.
\newblock \href {https://arxiv.org/abs/2311.04205} {Rephrase and respond: Let large language models ask better questions for themselves}.
\newblock \emph{Preprint}, arXiv:2311.04205.

\bibitem[{Eldan and Li(2023)}]{TinyStories}
Ronen Eldan and Yuanzhi Li. 2023.
\newblock \href {https://arxiv.org/abs/2305.07759} {Tinystories: How small can language models be and still speak coherent english?}
\newblock \emph{Preprint}, arXiv:2305.07759.

\bibitem[{Grattafiori et~al.(2024)Grattafiori, Dubey, and et~al.}]{llama3}
Aaron Grattafiori, Abhimanyu Dubey, and Abhinav~Jauhri et~al. 2024.
\newblock \href {https://arxiv.org/abs/2407.21783} {The llama 3 herd of models}.
\newblock \emph{Preprint}, arXiv:2407.21783.

\bibitem[{He et~al.(2024)He, Guo, and Zhu}]{replay-seekr}
Jinghan He, Haiyun Guo, and Kuan et~al. Zhu. 2024.
\newblock \href {https://doi.org/10.18653/v1/2024.emnlp-main.190} {{SEEKR}: Selective attention-guided knowledge retention for continual learning of large language models}.
\newblock In \emph{Proceedings of the 2024 Conference on Empirical Methods in Natural Language Processing}, pages 3254--3266, Miami, Florida, USA. Association for Computational Linguistics.

\bibitem[{He et~al.(2025)He, WANG, Weber, Zhu, Athiwaratkun, and Zhang}]{synthetic_data_1}
Linda He, Jue WANG, Maurice Weber, Shang Zhu, Ben Athiwaratkun, and Ce~Zhang. 2025.
\newblock \href {https://openreview.net/forum?id=BkwCrIsTbR} {Scaling instruction-tuned {LLM}s to million-token contexts via hierarchical synthetic data generation}.
\newblock In \emph{The Thirteenth International Conference on Learning Representations}.

\bibitem[{Hendrycks et~al.(2021)Hendrycks, Burns, Kadavath, Arora, Basart, Tang, Song, and Steinhardt}]{math500}
Dan Hendrycks, Collin Burns, Saurav Kadavath, Akul Arora, Steven Basart, Eric Tang, Dawn Song, and Jacob Steinhardt. 2021.
\newblock \href {https://openreview.net/forum?id=7Bywt2mQsCe} {Measuring mathematical problem solving with the {MATH} dataset}.
\newblock In \emph{Thirty-fifth Conference on Neural Information Processing Systems Datasets and Benchmarks Track (Round 2)}.

\bibitem[{Hu et~al.(2025)Hu, Peng, Bi, Shen, Liu, Lou, and Luo}]{jas_ours}
Zhenlin Hu, Zhizhi Peng, Zhen Bi, Qing Shen, Zhenfang Liu, Jungang Lou, and Xin Luo. 2025.
\newblock \href {https://doi.org/10.1109/JAS.2025.125540} {Advancing healthcare with large language models: Techniques and application}.
\newblock \emph{IEEE/CAA Journal of Automatica Sinica}, 12(12):2371--2398.

\bibitem[{Huang et~al.(2025{\natexlab{a}})Huang, Yu, Ma, Zhong, Feng, Wang, Chen, Peng, Feng, Qin, and Liu}]{Hallucination}
Lei Huang, Weijiang Yu, Weitao Ma, Weihong Zhong, Zhangyin Feng, Haotian Wang, Qianglong Chen, Weihua Peng, Xiaocheng Feng, Bing Qin, and Ting Liu. 2025{\natexlab{a}}.
\newblock \href {https://doi.org/10.1145/3703155} {A survey on hallucination in large language models: Principles, taxonomy, challenges, and open questions}.
\newblock \emph{ACM Transactions on Information Systems}, 43(2).

\bibitem[{Huang et~al.(2025{\natexlab{b}})Huang, Cheng, Zhang, Wang, Wang, Yan, and Wei}]{FAPM}
Wei Huang, Anda Cheng, Zhao Zhang, Yinggui Wang, Lei Wang, Shoumeng Yan, and Tao Wei. 2025{\natexlab{b}}.
\newblock \href {https://openreview.net/forum?id=fHvh913U1H} {Mitigating catastrophic forgetting in large language models with forgetting-aware pruning}.

\bibitem[{Liang et~al.(2026)Liang, Zhong, Xu, Jiang, Zhong, Fang, Gu, Deng, Yao, Wang, Qiao, Xu, Wu, Wang, Liu, Bi, Lou, Jiang, Zhu, Yu, Hong, Huang, Xue, Wang, Wang, Shan, Chen, Tu, Xiong, Xie, Zhang, Gui, Liang, Zhou, Wu, Shang, Gong, Lin, Xu, Deng, Zhang, Ding, Zhang, Huang, Zhang, Pan, Qi, Wang, and Chen}]{skill_net}
Yuan Liang, Ruobin Zhong, Haoming Xu, Chen Jiang, Yi~Zhong, Runnan Fang, Jia-Chen Gu, Shumin Deng, Yunzhi Yao, Mengru Wang, Shuofei Qiao, Xin Xu, Tongtong Wu, Kun Wang, Yang Liu, Zhen Bi, Jungang Lou, Yuchen~Eleanor Jiang, Hangcheng Zhu, and 30 others. 2026.
\newblock \href {https://arxiv.org/abs/2603.04448} {Skillnet: Create, evaluate, and connect ai skills}.
\newblock \emph{Preprint}, arXiv:2603.04448.

\bibitem[{Minegishi et~al.(2025)Minegishi, Furuta, Taniguchi, Iwasawa, and Matsuo}]{meta_learning_method1}
Gouki Minegishi, Hiroki Furuta, Shohei Taniguchi, Yusuke Iwasawa, and Yutaka Matsuo. 2025.
\newblock \href {https://openreview.net/forum?id=Xw01vF13aV} {Beyond induction heads: In-context meta learning induces multi-phase circuit emergence}.
\newblock In \emph{Forty-second International Conference on Machine Learning}.

\bibitem[{OpenAI(2024)}]{GPT-4}
OpenAI. 2024.
\newblock \href {https://arxiv.org/abs/2303.08774} {Gpt-4 technical report}.
\newblock \emph{Preprint}, arXiv:2303.08774.

\bibitem[{Prasad et~al.(2024)Prasad, Stengel-Eskin, and Bansal}]{synthetic_rephrase1}
Archiki Prasad, Elias Stengel-Eskin, and Mohit Bansal. 2024.
\newblock \href {https://openreview.net/forum?id=L4nOxziGf9} {Rephrase, augment, reason: Visual grounding of questions for vision-language models}.
\newblock In \emph{The Twelfth International Conference on Learning Representations}.

\bibitem[{Qwen et~al.(2025)Qwen, :, and et~al.}]{qwen2_5}
Qwen, :, and An~Yang et~al. 2025.
\newblock \href {https://arxiv.org/abs/2412.15115} {Qwen2.5 technical report}.
\newblock \emph{Preprint}, arXiv:2412.15115.

\bibitem[{Riaz et~al.(2025)Riaz, Bhabesh, Arannil, Ballesteros, and Horwood}]{MetaSynth}
Haris Riaz, Sourav Bhabesh, Vinayak Arannil, Miguel Ballesteros, and Graham Horwood. 2025.
\newblock \href {https://arxiv.org/abs/2504.12563} {Metasynth: Meta-prompting-driven agentic scaffolds for diverse synthetic data generation}.
\newblock \emph{Preprint}, arXiv:2504.12563.

\bibitem[{Shi et~al.(2025)Shi, Xu, Wang, Qin, Wang, Wang, Wang, Ebrahimi, and Wang}]{continual_learning_survery2}
Haizhou Shi, Zihao Xu, Hengyi Wang, Weiyi Qin, Wenyuan Wang, Yibin Wang, Zifeng Wang, Sayna Ebrahimi, and Hao Wang. 2025.
\newblock \href {https://doi.org/10.1145/3735633} {Continual learning of large language models: A comprehensive survey}.
\newblock \emph{ACM Comput. Surv.}, 58(5).

\bibitem[{Tong et~al.(2024)Tong, Zhang, Wang, Wu, and He}]{DART-Math}
Yuxuan Tong, Xiwen Zhang, Rui Wang, Ruidong Wu, and Junxian He. 2024.
\newblock \href {http://papers.nips.cc/paper\_files/paper/2024/hash/0ef1afa0daa888d695dcd5e9513bafa3-Abstract-Conference.html} {Dart-math: Difficulty-aware rejection tuning for mathematical problem-solving}.
\newblock In \emph{Advances in Neural Information Processing Systems 38: Annual Conference on Neural Information Processing Systems 2024, NeurIPS 2024, Vancouver, BC, Canada, December 10 - 15, 2024}.

\bibitem[{Wang et~al.(2018)Wang, Singh, Michael, Hill, Levy, and Bowman}]{glue-mnli}
Alex Wang, Amanpreet Singh, Julian Michael, Felix Hill, Omer Levy, and Samuel Bowman. 2018.
\newblock \href {https://doi.org/10.18653/v1/W18-5446} {{GLUE}: A multi-task benchmark and analysis platform for natural language understanding}.
\newblock In \emph{Proceedings of the 2018 {EMNLP} Workshop {B}lackbox{NLP}: Analyzing and Interpreting Neural Networks for {NLP}}, pages 353--355, Brussels, Belgium. Association for Computational Linguistics.

\bibitem[{Wang et~al.(2025{\natexlab{a}})Wang, Lyu, Sun, and Jing}]{GORP}
Chenxu Wang, Yilin Lyu, Zicheng Sun, and Liping Jing. 2025{\natexlab{a}}.
\newblock \href {https://doi.org/10.18653/v1/2025.acl-long.721} {Continual gradient low-rank projection fine-tuning for {LLM}s}.
\newblock In \emph{Proceedings of the 63rd Annual Meeting of the Association for Computational Linguistics (Volume 1: Long Papers)}, pages 14815--14829, Vienna, Austria. Association for Computational Linguistics.

\bibitem[{Wang et~al.(2025{\natexlab{b}})Wang, Chandra, and Zhang}]{continual_learning_method1}
Jiuqi Wang, Rohan Chandra, and Shangtong Zhang. 2025{\natexlab{b}}.
\newblock \href {https://arxiv.org/abs/2503.20018} {Experience replay addresses loss of plasticity in continual learning}.
\newblock \emph{Preprint}, arXiv:2503.20018.

\bibitem[{Wang et~al.(2024{\natexlab{a}})Wang, Zhang, Su, and Zhu}]{continual_learning_survery1}
Liyuan Wang, Xingxing Zhang, Hang Su, and Jun Zhu. 2024{\natexlab{a}}.
\newblock \href {https://doi.org/10.1109/TPAMI.2024.3367329} {A comprehensive survey of continual learning: Theory, method and application}.
\newblock \emph{IEEE Transactions on Pattern Analysis and Machine Intelligence}, 46(8):5362--5383.

\bibitem[{Wang et~al.(2025{\natexlab{c}})Wang, Zhao, Guo, and Wang}]{meta_learning_method2}
Maolin Wang, Xiangyu Zhao, Ruocheng Guo, and Junhui Wang. 2025{\natexlab{c}}.
\newblock \href {https://doi.org/10.1109/ICDE65448.2025.00376} {Metalora: Tensor-enhanced adaptive low-rank fine-tuning}.
\newblock In \emph{2025 IEEE 41st International Conference on Data Engineering (ICDE)}, pages 4680--4684.

\bibitem[{Wang et~al.(2025{\natexlab{d}})Wang, Tissue, Wang, Li, and Zeng}]{continual_learning_method}
Xingjin Wang, Howe Tissue, Lu~Wang, Linjing Li, and Daniel~Dajun Zeng. 2025{\natexlab{d}}.
\newblock \href {https://openreview.net/forum?id=Vk1rNMl0J1} {Learning dynamics in continual pre-training for large language models}.
\newblock In \emph{Forty-second International Conference on Machine Learning}.

\bibitem[{Wang et~al.(2024{\natexlab{b}})Wang, Liu, Shi et~al.}]{replay-inscl}
Yifan Wang, Yafei Liu, Chufan Shi, and 1 others. 2024{\natexlab{b}}.
\newblock \href {https://doi.org/10.18653/v1/2024.naacl-long.37} {{I}ns{CL}: A data-efficient continual learning paradigm for fine-tuning large language models with instructions}.
\newblock In \emph{Proceedings of the 2024 Conference of the North American Chapter of the Association for Computational Linguistics: Human Language Technologies (Volume 1: Long Papers)}, pages 663--677, Mexico City, Mexico. Association for Computational Linguistics.

\bibitem[{Wang et~al.(2023)Wang, Kordi, Mishra, Liu, Smith, Khashabi, and Hajishirzi}]{Self-Instruct}
Yizhong Wang, Yeganeh Kordi, Swaroop Mishra, Alisa Liu, Noah~A. Smith, Daniel Khashabi, and Hannaneh Hajishirzi. 2023.
\newblock \href {https://doi.org/10.18653/v1/2023.acl-long.754} {Self-instruct: Aligning language models with self-generated instructions}.
\newblock In \emph{Proceedings of the 61st Annual Meeting of the Association for Computational Linguistics (Volume 1: Long Papers)}, pages 13484--13508, Toronto, Canada. Association for Computational Linguistics.

\bibitem[{Xiao et~al.(2025)Xiao, Zeng, Chen, Li, Bertsch, and Neubig}]{meta_learning_method3}
Emily Xiao, Yixiao Zeng, Ada Chen, Chin-Jou Li, Amanda Bertsch, and Graham Neubig. 2025.
\newblock \href {https://arxiv.org/abs/2510.16932} {Prompt-mii: Meta-learning instruction induction for llms}.
\newblock \emph{Preprint}, arXiv:2510.16932.

\bibitem[{Xu et~al.(2024)Xu, Sun, Zheng, Geng, Zhao, Feng, Tao, Lin, and Jiang}]{WizardLM}
Can Xu, Qingfeng Sun, Kai Zheng, Xiubo Geng, Pu~Zhao, Jiazhan Feng, Chongyang Tao, Qingwei Lin, and Daxin Jiang. 2024.
\newblock \href {https://openreview.net/forum?id=CfXh93NDgH} {Wizard{LM}: Empowering large pre-trained language models to follow complex instructions}.
\newblock In \emph{The Twelfth International Conference on Learning Representations}.

\bibitem[{Xue et~al.(2026{\natexlab{a}})Xue, Liao, Qin, Zhang, Mu, Zhou, and Yu}]{38_xue_2026coreb}
Siqiao Xue, Zihan Liao, Jin Qin, Ziyin Zhang, Yixiang Mu, Fan Zhou, and Hang Yu. 2026{\natexlab{a}}.
\newblock \href {https://arxiv.org/abs/2605.04615} {Beyond retrieval: A multitask benchmark and model for code search}.
\newblock \emph{arXiv preprint arXiv:2605.04615}.

\bibitem[{Xue et~al.(2026{\natexlab{b}})Xue, Zhu, Zhang, Cai, Wang, Mu, Zhou, Li, Di, and Yu}]{37_xue_2026quitobench}
Siqiao Xue, Zhaoyang Zhu, Wei Zhang, Rongyao Cai, Rui Wang, Yixiang Mu, Fan Zhou, Jianguo Li, Peng Di, and Hang Yu. 2026{\natexlab{b}}.
\newblock \href {https://arxiv.org/abs/2603.26017} {Quitobench: A high-quality open time series forecasting benchmark}.
\newblock \emph{arXiv preprint arXiv:2603.26017}.

\bibitem[{Xue et~al.(2026{\natexlab{c}})Xue, Bi, Ma, Hu, Wang, Chen, Liu, Zhao, Xiao, and Lou}]{zihao_xue_tp}
Zihao Xue, Zhen Bi, Long Ma, Zhenlin Hu, Yan Wang, Xueshu Chen, Zhenfang Liu, Kang Zhao, Jie Xiao, and Jungang Lou. 2026{\natexlab{c}}.
\newblock \href {https://arxiv.org/abs/2507.12314} {Thought purity: A defense framework for chain-of-thought attack}.
\newblock \emph{Preprint}, arXiv:2507.12314.

\bibitem[{Yin et~al.(2025)Yin, Wei, Zhu, Chen, and Meng}]{synthetic_1}
Shangjian Yin, Zhepei Wei, Xinyu Zhu, Wei-Lin Chen, and Yu~Meng. 2025.
\newblock \href {https://arxiv.org/abs/2510.06652} {Aligning large language models via fully self-synthetic data}.
\newblock \emph{Preprint}, arXiv:2510.06652.

\bibitem[{Yu et~al.(2024)Yu, Jiang, Shi, YU, Liu, Zhang, Kwok, Li, Weller, and Liu}]{synthetic_metamath}
Longhui Yu, Weisen Jiang, Han Shi, Jincheng YU, Zhengying Liu, Yu~Zhang, James Kwok, Zhenguo Li, Adrian Weller, and Weiyang Liu. 2024.
\newblock \href {https://openreview.net/forum?id=N8N0hgNDRt} {Metamath: Bootstrap your own mathematical questions for large language models}.
\newblock In \emph{The Twelfth International Conference on Learning Representations}.

\bibitem[{Yu et~al.(2023{\natexlab{a}})Yu, Zhuang, Zhang, Meng, Ratner, Krishna, Shen, and Zhang}]{attroprompt}
Yue Yu, Yuchen Zhuang, Jieyu Zhang, Yu~Meng, Alexander Ratner, Ranjay Krishna, Jiaming Shen, and Chao Zhang. 2023{\natexlab{a}}.
\newblock \href {https://arxiv.org/abs/2306.15895} {Large language model as attributed training data generator: A tale of diversity and bias}.
\newblock \emph{Preprint}, arXiv:2306.15895.

\bibitem[{Yu et~al.(2023{\natexlab{b}})Yu, Zhuang, Zhang, Meng, Ratner, Krishna, Shen, and Zhang}]{synthetic_prompt_1}
Yue Yu, Yuchen Zhuang, Jieyu Zhang, Yu~Meng, Alexander~J. Ratner, Ranjay Krishna, Jiaming Shen, and Chao Zhang. 2023{\natexlab{b}}.
\newblock \href {http://papers.nips.cc/paper\_files/paper/2023/hash/ae9500c4f5607caf2eff033c67daa9d7-Abstract-Datasets\_and\_Benchmarks.html} {Large language model as attributed training data generator: {A} tale of diversity and bias}.
\newblock In \emph{Advances in Neural Information Processing Systems 36: Annual Conference on Neural Information Processing Systems 2023, NeurIPS 2023, New Orleans, LA, USA, December 10 - 16, 2023}.

\bibitem[{Zeng et~al.(2024)Zeng, Xu, Zhao, Lou, and Chen}]{Automatic_Instruction}
Weihao Zeng, Can Xu, Yingxiu Zhao, Jian-Guang Lou, and Weizhu Chen. 2024.
\newblock \href {https://doi.org/10.18653/v1/2024.emnlp-main.397} {Automatic instruction evolving for large language models}.
\newblock In \emph{Proceedings of the 2024 Conference on Empirical Methods in Natural Language Processing}, pages 6998--7018, Miami, Florida, USA. Association for Computational Linguistics.

\bibitem[{Zhan et~al.(2025)Zhan, Lai, Lu, Lin, Yang, and Tan}]{MathSmith}
Shaoxiong Zhan, Yanlin Lai, Ziyu Lu, Dahua Lin, Ziqing Yang, and Fei Tan. 2025.
\newblock \href {https://arxiv.org/abs/2508.05592} {Mathsmith: Towards extremely hard mathematical reasoning by forging synthetic problems with a reinforced policy}.
\newblock \emph{Preprint}, arXiv:2508.05592.

\bibitem[{Zhang et~al.(2015)Zhang, Zhao, and LeCun}]{stand_cl_3_dataset}
Xiang Zhang, Junbo~Jake Zhao, and Yann LeCun. 2015.
\newblock \href {http://papers.nips.cc/paper/5782-character-level-convolutional-networks-for-text-classification} {Character-level convolutional networks for text classification}.
\newblock In \emph{NIPS}, pages 649--657.

\end{thebibliography}



\appendix

\newpage

\section{Additional Discussion of Related Paradigms}
\label{app:paradigm_discussion}

To complement the comparison in Table~\ref{tab:paradigm_comparison}, we provide a more detailed explanation of how \textbf{StreamSynth} differs from several related paradigms.

\paragraph{Static synthesis.}
Static synthesis methods generate synthetic data for a specific task using prompts, templates, seed examples, rewriting rules, or heuristic evolution strategies. 
Although these methods can improve data diversity and downstream performance, they typically treat each synthesis task independently. 
The synthesis process does not explicitly accumulate experience from previous tasks, nor is the model optimized to improve future synthesis behavior over a task stream.

\paragraph{Multi-task synthesis.}
Multi-task synthesis extends static synthesis by exploiting shared information across multiple observed tasks or domains. 
For example, a synthesis system may jointly use examples, attributes, or prompts from several datasets to improve coverage. 
This setting can benefit from cross-task signals, which is why we mark it as using task experience in Table~\ref{tab:paradigm_comparison}. 
However, multi-task synthesis usually assumes that multiple tasks are jointly available during construction, rather than arriving sequentially under a strict streaming protocol. 
Therefore, it does not directly study how synthesis behavior should be updated when future tasks are unavailable.

\paragraph{Continual learning.}
Continual learning focuses on learning from sequentially arriving tasks while preserving predictive performance on previous tasks. 
Typical objectives include mitigating catastrophic forgetting, retaining task knowledge, or improving adaptation under distribution shifts. 
Although these goals are related to StreamSynth, the primary target of continual learning is usually the downstream predictor rather than the synthesis process itself. 
In contrast, StreamSynth focuses on whether a synthesis model can acquire reusable generation behavior from historical synthesis experience and transfer it to future data construction tasks.

\paragraph{Meta-learning.}
Meta-learning aims to acquire task-level priors that enable fast adaptation to new tasks. 
It often assumes access to a distribution of tasks or meta-training episodes, and the learned initialization or adaptation rule is optimized for future task adaptation. 
StreamSynth differs in that tasks arrive sequentially, future tasks are not available during earlier stages, and the objective is not merely fast adaptation but learning transferable synthesis behavior through feedback over the task stream.

\paragraph{StreamSynth.}
StreamSynth combines several requirements that are not jointly addressed by the above paradigms: sequential task arrival, a synthesis-oriented objective, the use of historical synthesis experience, feedback-driven optimization, and generalization to future synthesis tasks. 
This distinction is the motivation for formulating StreamSynth as a separate setting rather than treating it as a direct instance of existing continual learning or meta-learning frameworks.

\section{Additional Method Details}
\label{app:method_details}

\paragraph{Details of reward components.}
In the main text, we summarize the sample-level reward using structural validity, fluency, and task relevance. 
Structural validity is implemented as a rule-based check over output format, required fields, and malformed expressions. 
Fluency combines normalized language-model likelihood and lightweight style statistics, such as sentence length, punctuation balance, and surface coherence. 
Task relevance is computed using task-specific classifiers and keyword-based matching signals, where the classifier and keyword set are instantiated according to the current task.

\paragraph{Details of set-level diversity.}
For set-level scoring, each generated sample is compared with samples in the same mini-batch using sentence embeddings.
The proximity weights are obtained by normalizing pairwise similarities, so closer neighbors contribute more to the local density estimate.
The distinctiveness score then penalizes samples located in crowded semantic regions, encouraging the model to generate diverse but task-compatible samples.

\section{Experimental Details}
\label{appendix:exp_details}

This appendix provides detailed descriptions of datasets, task stream construction, model configurations, baseline methods, and evaluation protocols used in our experiments. The anonymized code for this work is publicly available at: 
\url{https://github.com/a18538308316-bot/StreamSynth}

\subsection{Datasets and Task Stream}

We conduct experiments on four widely used real-world NLP benchmarks: \textbf{Yelp}, \textbf{Yahoo}, and \textbf{Amazon} from the \textbf{StandardCL} benchmark suite, as well as \textbf{MNLI} from the GLUE benchmark.
These datasets cover diverse language understanding scenarios, including review understanding, topic-oriented text understanding, and natural language inference, enabling us to evaluate synthesis transfer across different task types and domains.

For the main experiments, we use \textbf{Yelp} $\rightarrow$ \textbf{Amazon} $\rightarrow$ \textbf{Yahoo} $\rightarrow$ \textbf{MNLI} as the default task stream.
This stream is designed to include both relatively close transfer settings and more challenging cross-task transfer settings, moving from review-oriented understanding to broader text understanding and natural language inference.
At each stage, the synthesis model only has access to the current task, while future tasks remain unseen.
Importantly, this default order is used for the main comparison only; we further evaluate alternative task permutations in the ablation study to examine order robustness.

For each task, due to the large number of available samples, we randomly select \textbf{4,000} instances for training and \textbf{7,600} instances for testing.
Synthetic samples are generated for the training split only, while the original test sets are kept unchanged to ensure fair and consistent evaluation across all methods.

\subsection{Models and Training Protocol}

We adopt two instruction-tuned large language models as the backbone for both synthesis and downstream evaluation: \textbf{LLaMA3.1-8B} and \textbf{Qwen2.5-7B}.
In each experiment, the same model architecture is used to generate synthetic data and to serve as the downstream model trained on the augmented dataset.
This design avoids potential confounding effects caused by model mismatch and ensures that performance differences mainly reflect the effectiveness and transferability of synthesized data.

For downstream training, we construct mixed training sets by augmenting the original data with an equal number of synthetic samples generated by each method.
All downstream models are trained using identical optimization settings and evaluated on the unchanged test splits of each dataset.

\subsection{Baseline Methods}

We compare \textbf{SynLearner} with a diverse set of baselines covering prompt-based direct synthesis, parameter-efficient continual adaptation, and replay- or retention-based adaptation methods.
These baselines are selected to evaluate whether SynLearner can learn transferable synthesis behavior beyond standard prompt engineering or conventional sequential adaptation.

\paragraph{Direct Synthesis (No Training).}
These methods generate synthetic data by prompting pretrained LLMs without updating model parameters.
Importantly, ``no training'' does not imply naive prompting: several strong baselines use class-conditional, attribute-conditioned, or structured prompts to improve controllability and diversity.

Specifically, we consider:
\begin{itemize}[nolistsep, leftmargin=*]
    \item \textbf{Ori}: A lower-bound baseline where the downstream model is trained solely on the original dataset without synthetic augmentation.
    
    \item \textbf{LLaMA3.1-8B / LLaMA3.1-70B}~\citep{llama3}: Synthetic samples are generated by LLaMA3.1 models through class-conditional or attribute-based prompting, allowing us to evaluate prompt-based synthesis under different model scales.
    
    \item \textbf{Qwen2.5-7B}~\citep{qwen2_5}: An instruction-tuned open-source LLM used for direct prompt-based synthesis under the same prompting protocol.
    
    \item \textbf{DeepSeek}~\citep{Deepseek}: A strong zero-shot synthesis baseline that generates synthetic samples without task-specific adaptation.
\end{itemize}

Some prompt-based baselines adopt attributed prompting techniques such as \textbf{AttrPrompt}~\citep{attroprompt}, which condition generation on multiple attributes such as style, polarity, length, or task-specific constraints.
This ensures that the direct synthesis baselines are not limited to simple label-conditioned prompting, but also include stronger prompt-engineered generation strategies.

\paragraph{Parameter-Efficient Continual Adaptation.}
This group contains methods that adapt model parameters across sequential tasks while attempting to reduce interference between old and new tasks.
Unlike SynLearner, these methods are originally designed for maintaining or adapting model behavior under task streams, rather than explicitly learning reusable synthesis strategies.

We include:
\begin{itemize}[nolistsep, leftmargin=*]
    \item \textbf{GORP}~\citep{GORP}: A gradient-projection-based method that constrains parameter updates within orthogonal low-rank subspaces, reducing interference from newly arriving tasks.
    
    \item \textbf{FAPM}~\citep{FAPM}: A Fisher-based adaptive parameter masking method that identifies and preserves important parameters to mitigate forgetting during sequential adaptation.
\end{itemize}

These baselines test whether conventional parameter-efficient adaptation mechanisms can support synthesis learning under the StreamSynth setting.

\paragraph{Replay- and Retention-Based Adaptation.}
This group focuses on preserving previously acquired task knowledge through replay selection or selective retention mechanisms.
Such methods are relevant to StreamSynth because they also operate under sequential task arrival, but their primary objective is knowledge retention rather than feedback-driven synthesis improvement.

We include:
\begin{itemize}[nolistsep, leftmargin=*]
    \item \textbf{InsCL}~\citep{replay-inscl}: A replay-based continual learning method that selects representative instruction data based on instruction-level task similarity, helping preserve instruction-following ability across tasks.
    
    \item \textbf{SEEKR}~\citep{replay-seekr}: A selective retention method that identifies task-relevant attention components and preserves important attention patterns to maintain performance under task streams.
\end{itemize}

Together, these three groups provide complementary comparisons: direct synthesis evaluates the strength of prompt-only generation, parameter-efficient continual adaptation evaluates sequential parameter updating, and replay- or retention-based adaptation evaluates the role of preserving historical task knowledge.
This allows us to assess whether SynLearner provides additional benefits by explicitly learning synthesis behavior from feedback and accumulated experience.

\subsection{Evaluation Protocol}

For all methods, downstream evaluation is conducted by training the model on the mixed dataset (original + synthetic samples) and reporting performance on the original test sets.
This protocol ensures that improvements are attributable to the quality and transferability of synthesized data rather than changes to evaluation data.

All experiments follow the same training and evaluation procedures across methods, enabling fair comparison under the StreamSynth setting.

\subsection{Computational Resources and Budget}

All experiments are conducted on NVIDIA A800 GPUs.
For each task in the StreamSynth pipeline, the Hierarchical Reward Optimization (HRO) stage requires approximately \textbf{120 GPU hours} for training.
The total computational cost scales linearly with the number of tasks in the stream.

The backbone models used in our experiments include \textbf{LLaMA3.1-8B} and \textbf{Qwen2.5-7B}, whose parameter sizes are reported in their respective original papers.
No additional large-scale pretraining is performed; all updates are limited to task-level fine-tuning and reinforcement learning under the StreamSynth setting.

\subsection{Hyperparameters and Experimental Stability}

Hyperparameters for supervised fine-tuning and hierarchical reward optimization are selected through empirical validation.
We perform controlled hyperparameter exploration to identify stable and effective configurations, and the final settings are fixed across all experiments for fair comparison.
Detailed hyperparameter choices and training configurations are provided in the released codebase.

To reduce variance, all reported results are obtained by repeating experiments multiple times with different random seeds and reporting the averaged performance.
This protocol ensures that observed improvements are stable and not attributable to randomness in initialization or data sampling.

\begin{table*}[ht!]
\centering
\footnotesize
\resizebox{\textwidth}{!}{
\begin{tabular}{llcccccccc}
\toprule
\multirow{2}{*}{\textbf{Model}} & \multirow{2}{*}{\textbf{Methods}} & \multicolumn{2}{c}{\fontsize{8pt}{9pt}\selectfont \textbf{{Step 1:}} Yelp}  
& \multicolumn{2}{c}{\fontsize{8pt}{9pt}\selectfont \textbf{{Step 2:}} Amazon}
& \multicolumn{2}{c}{\fontsize{8pt}{9pt}\selectfont \textbf{{Step 3:}} Yahoo}
& \multicolumn{2}{c}{\fontsize{8pt}{9pt}\selectfont \textbf{{Step 4:}} MNLI} \\
\cmidrule(lr){3-4} \cmidrule(lr){5-6} \cmidrule(lr){7-8} \cmidrule(lr){9-10} 
&
& \multicolumn{1}{c}{\fontsize{5pt}{6pt}\selectfont Acc (\%)}   
& \multicolumn{1}{c}{\fontsize{5pt}{6pt}\selectfont F1-Scores (\%)}  
& \multicolumn{1}{c}{\fontsize{5pt}{6pt}\selectfont Acc (\%)}   
& \multicolumn{1}{c}{\fontsize{5pt}{6pt}\selectfont F1-Scores (\%)}  
& \multicolumn{1}{c}{\fontsize{5pt}{6pt}\selectfont Acc (\%)}   
& \multicolumn{1}{c}{\fontsize{5pt}{6pt}\selectfont F1-Scores (\%)}   
& \multicolumn{1}{c}{\fontsize{5pt}{6pt}\selectfont Acc (\%)}   
& \multicolumn{1}{c}{\fontsize{5pt}{6pt}\selectfont F1-Scores (\%)} \\
\midrule
\multirow{3}{*}{\scriptsize LLaMA} 
& \scriptsize Ours-LLaMA3.2-1B & \scriptsize 38.53 & \scriptsize 34.94 & \scriptsize 36.88 & \scriptsize 33.78 & \scriptsize 45.42  & \scriptsize 45.71 & \scriptsize 35.52 & \scriptsize 35.49  \\
& \scriptsize Ours-LLaMA3.1-8B & \scriptsize 66.11 & \scriptsize 65.76 & \scriptsize 60.09  & \scriptsize 60.39  & \scriptsize 70.34  & \scriptsize 70.11 & \scriptsize 82.50  & \scriptsize 82.28   \\
& \scriptsize Ours-LLaMA3.1-13B  & \scriptsize 67.39 & \scriptsize 67.00 & \scriptsize 60.13 & \scriptsize 60.19 & \scriptsize 71.20 & \scriptsize 70.74 & \scriptsize 83.80 & \scriptsize 83.74 \\
\midrule  
\multirow{3}{*}{\scriptsize Qwen} 
& \scriptsize Ours-Qwen2.5-1.5B & \scriptsize 56.25 & \scriptsize 56.53 & \scriptsize 49.11 & \scriptsize 49.13 & \scriptsize 55.00 & \scriptsize 56.70  & \scriptsize 73.70 & \scriptsize 73.09 \\
& \scriptsize Ours-Qwen2.5-7B & \scriptsize 63.91  & \scriptsize 64.13 & \scriptsize 58.59  & \scriptsize 58.66  & \scriptsize 69.25 & \scriptsize 68.94 & \scriptsize 86.56   & \scriptsize 86.47  \\
& \scriptsize Ours-Qwen2.5-14B & \scriptsize 64.30 & \scriptsize 64.78 & \scriptsize 58.99 & \scriptsize 59.19 & \scriptsize 71.11 & \scriptsize 70.94 & \scriptsize 87.99 & \scriptsize 87.88 \\
\bottomrule
\end{tabular}}
\caption{Ablation with Different Parameter Quantities.}
\label{tab:scaling_law}
\end{table*}

\begin{table*}[ht!]
\centering
\footnotesize
\resizebox{\textwidth}{!}{
\begin{tabular}{llcccccccc}
\toprule

\multirow{2}{*}{\textbf{Model}} & \multirow{2}{*}{\textbf{Methods}} 
& \multicolumn{2}{c}{\fontsize{8pt}{9pt}\selectfont \textbf{{Step 1:}} Yelp}  
& \multicolumn{2}{c}{\fontsize{8pt}{9pt}\selectfont \textbf{{Step 2:}} Amazon}
& \multicolumn{2}{c}{\fontsize{8pt}{9pt}\selectfont \textbf{{Step 3:}} Yahoo}
& \multicolumn{2}{c}{\fontsize{8pt}{9pt}\selectfont \textbf{{Step 4:}} MNLI}
\\
\cmidrule(lr){3-4} \cmidrule(lr){5-6} \cmidrule(lr){7-8} \cmidrule(lr){9-10} 
&
& \multicolumn{1}{c}{\fontsize{5pt}{6pt}\selectfont Acc (\%)}   
& \multicolumn{1}{c}{\fontsize{5pt}{6pt}\selectfont F1-Scores (\%)}  
& \multicolumn{1}{c}{\fontsize{5pt}{6pt}\selectfont Acc (\%)}   
& \multicolumn{1}{c}{\fontsize{5pt}{6pt}\selectfont F1-Scores (\%)}  
& \multicolumn{1}{c}{\fontsize{5pt}{6pt}\selectfont Acc (\%)}   
& \multicolumn{1}{c}{\fontsize{5pt}{6pt}\selectfont F1-Scores (\%)}   
& \multicolumn{1}{c}{\fontsize{5pt}{6pt}\selectfont Acc (\%)}   
& \multicolumn{1}{c}{\fontsize{5pt}{6pt}\selectfont F1-Scores (\%)}  
\\

\midrule

\multirow{17}{*}{\scriptsize LLaMA3.1-8B} 





& \scriptsize FAPM-Order1 & \scriptsize 64.38  & \scriptsize 63.70 & \scriptsize 57.66  & \scriptsize 57.26  & \scriptsize 66.01  & \scriptsize 66.24 & \scriptsize 79.21  & \scriptsize 79.20  \\

& \scriptsize GORP-Order1 & \scriptsize 64.43  & \scriptsize 64.75 & \scriptsize 57.95  & \scriptsize 58.08  & \scriptsize 64.41  & \scriptsize 64.61 & \scriptsize 79.74  & \scriptsize 79.38  \\

& \scriptsize InsCL-Order1 & \scriptsize 64.05  & \scriptsize 63.83 & \scriptsize 58.17  & \scriptsize 58.32  & \scriptsize 67.34  & \scriptsize 66.93 & \scriptsize 78.71  & \scriptsize 78.77  \\

& \scriptsize SEEKR-Order1 & \scriptsize 64.24  & \scriptsize 64.57 & \scriptsize 58.41  & \scriptsize 58.37  & \scriptsize 66.71  & \scriptsize 66.27 & \scriptsize 80.68  & \scriptsize 80.41  \\

& \scriptsize Ours-Order1 & \scriptsize 66.68  & \scriptsize 66.46 & \scriptsize 58.46  & \scriptsize 58.48  & \scriptsize 67.05  & \scriptsize 67.41 & \scriptsize 81.71  & \scriptsize 81.21  \\

\cmidrule(lr){2-10}

& \scriptsize FAPM-Order2 & \scriptsize 65.64  & \scriptsize 65.25 & \scriptsize 58.76  & \scriptsize 58.78  & \scriptsize 64.12  & \scriptsize 64.43 & \scriptsize 79.90  & \scriptsize 79.84  \\

& \scriptsize GORP-Order2 & \scriptsize 65.24  & \scriptsize 64.94 & \scriptsize 59.08  & \scriptsize 59.13  & \scriptsize 66.13  & \scriptsize 66.49 & \scriptsize 80.76  & \scriptsize 80.48  \\

& \scriptsize InsCL-Order2 & \scriptsize 63.20  & \scriptsize 63.06 & \scriptsize 58.28  & \scriptsize 58.61  & \scriptsize 62.80  & \scriptsize 63.18 & \scriptsize 79.43 & \scriptsize 79.10  \\

& \scriptsize SEEKR-Order2 & \scriptsize 64.72  & \scriptsize 64.36 & \scriptsize 58.25  & \scriptsize 58.51  & \scriptsize 66.11  & \scriptsize 66.70 & \scriptsize 78.84  & \scriptsize 78.97  \\

& \scriptsize Ours-Order2 & \scriptsize 67.11  & \scriptsize 67.22 & \scriptsize 58.63  & \scriptsize 58.46  & \scriptsize 69.36  & \scriptsize 69.21 & \scriptsize 81.04  & \scriptsize 80.81  \\

\cmidrule(lr){2-10}

& \scriptsize FAPM-Order3 & \scriptsize 65.37  & \scriptsize 65.47 & \scriptsize 57.78  & \scriptsize 57.56  & \scriptsize 63.96  & \scriptsize 64.79 & \scriptsize 80.31  & \scriptsize 79.66  \\

& \scriptsize GORP-Order3 & \scriptsize 65.84  & \scriptsize 64.95 & \scriptsize 57.88  & \scriptsize 58.06  & \scriptsize 65.64  & \scriptsize 66.03 & \scriptsize 80.37  & \scriptsize 80.24  \\

& \scriptsize InsCL-Order3 & \scriptsize 63.67  & \scriptsize 62.83 & \scriptsize 58.96  & \scriptsize 59.17  & \scriptsize 66.39  & \scriptsize 66.64 & \scriptsize 79.21 & \scriptsize 78.37  \\

& \scriptsize SEEKR-Order3 & \scriptsize 64.26  & \scriptsize 63.61 & \scriptsize 57.91  & \scriptsize 57.81  & \scriptsize 64.92  & \scriptsize 65.20 & \scriptsize 78.93  & \scriptsize 78.79  \\

& \scriptsize Ours-Order3 & \scriptsize 67.17  & \scriptsize 67.22 & \scriptsize 59.21  & \scriptsize 59.29  & \scriptsize 67.00  & \scriptsize 67.52 & \scriptsize 81.08  & \scriptsize 80.64  \\

\bottomrule
\end{tabular}}
\caption{
Task order ablation of \textbf{SynLearner} under the \textbf{StreamSynth} setting on LLaMA3.1-8B.
“Order1–3” correspond to different task stream permutations:
Order1 (Amazon $\rightarrow$ MNLI $\rightarrow$ Yahoo $\rightarrow$ Yelp),
Order2 (MNLI $\rightarrow$ Amazon $\rightarrow$ Yelp $\rightarrow$ Yahoo),
and Order3 (MNLI $\rightarrow$ Yahoo $\rightarrow$ Amazon $\rightarrow$ Yelp).
Results show that \textbf{SynLearner} maintains strong and consistent performance across all task orders,
indicating robustness to task stream permutations rather than reliance on a specific ordering.
}
\label{tab:ablation}
\end{table*}

\section{Additional Experiment Results}
\label{app:extra_exp}

\paragraph{Task Order Robustness and Performance Consistency.}

To complement the visualization results in Figure~\ref{fig:task_order}, 
Table~\ref{tab:ablation} reports the full quantitative performance of different methods under three task stream permutations.
Each block corresponds to a distinct task order, and all methods are evaluated using the same backbone and evaluation protocol.

Across all three task orders, \textbf{SynLearner} consistently achieves strong performance on every dataset in the stream, 
and maintains its relative advantage regardless of how tasks are permuted.
Importantly, the performance gains of SynLearner are observed across all orders rather than being tied to a specific task sequence,
indicating that its effectiveness does not rely on a carefully engineered curriculum.

This order-agnostic advantage suggests that SynLearner learns general synthesis strategies that transfer across tasks,
instead of overfitting to particular task arrangements.
As a result, the framework remains effective even when the task stream is re-ordered,
which is critical for realistic streaming synthesis scenarios where task arrival order cannot be controlled.

\paragraph{Detailed Analysis of Task Position Effects.}

Beyond overall task order robustness, Table~\ref{tab:ablation} also reveals a consistent position-dependent performance pattern for SynLearner.
When a dataset appears later in the task stream, SynLearner generally achieves higher synthesis accuracy and F1 scores compared to when the same dataset is placed earlier.
This trend is particularly evident on complex tasks such as MNLI, where performance improvements of over 1.0 point are observed when the task is positioned in later stages.

In contrast, baseline methods do not exhibit a clear or consistent relationship between task position and synthesis performance.
Their results often remain flat or fluctuate irregularly across different stream positions, indicating limited ability to benefit from prior synthesis experience.

These observations align with the findings in Figure~\ref{fig:order_analyze} and provide additional numerical support that SynLearner effectively accumulates synthesis knowledge over the task stream.
Rather than treating each task independently, the model leverages historical synthesis experience to improve performance on subsequent tasks.

\begin{figure*}[t]
  \centering
  \includegraphics[width=1.0\linewidth]{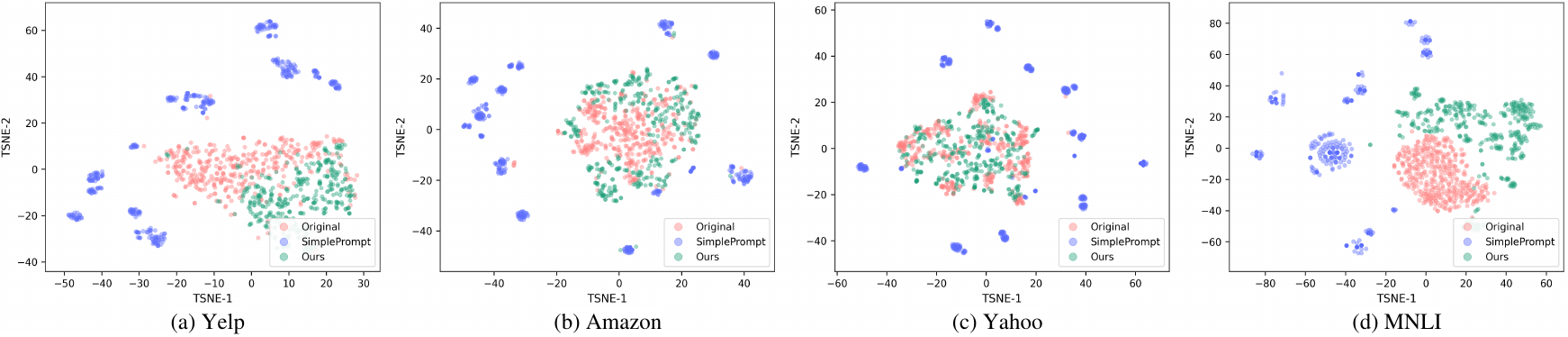}
  \caption{
  Full t-SNE visualization of original and synthetic samples across four tasks.
  \textit{SimplePrompt} denotes a basic few-shot prompting strategy without dynamic or evolutionary prompt expansion.
  Compared with SimplePrompt, SynLearner generally provides broader semantic coverage while preserving task-relevant alignment with the original data distribution.
  }
  \label{fig:tsne_distribution_full}
\end{figure*}

\paragraph{Qualitative Analysis of Synthetic Data Distributions.}

To complement the component ablation results, we provide full t-SNE visualizations of synthesized samples in Figure~\ref{fig:tsne_distribution_full}.
We compare three sources of data: original training samples, samples generated by a simple few-shot prompting baseline, and samples generated by SynLearner.
The simple prompting baseline directly asks the model to synthesize task-specific samples using basic task instructions and few-shot examples, without dynamic slot construction or evolutionary prompt expansion.

Overall, SimplePrompt often produces clustered or scattered samples that either occupy narrow regions or deviate from the main data manifold.
In contrast, SynLearner tends to cover broader semantic regions while maintaining meaningful proximity to the original distribution.
This indicates that the proposed diversity-aware prompting strategy does not merely increase random variation, but helps construct a richer and more task-compatible synthesis space.

The patterns vary across datasets.
On Yelp, SynLearner shows some distance from the original distribution, but it also expands the semantic coverage beyond the original samples, providing complementary diversity.
On Amazon and Yahoo, SynLearner maintains stronger overlap with the original distribution while still expanding the coverage of synthesized samples, suggesting a better balance between diversity and distributional alignment.
On MNLI, the distribution shift is more visible, likely because NLI examples involve sentence-pair relations and more complex semantic structures than single-text understanding tasks.
Nevertheless, the expanded coverage on MNLI still supports the role of SynLearner in improving synthesis diversity, which is consistent with its downstream performance gains.

\begin{figure*}[t]
  \centering
  \includegraphics[width=0.98\linewidth]{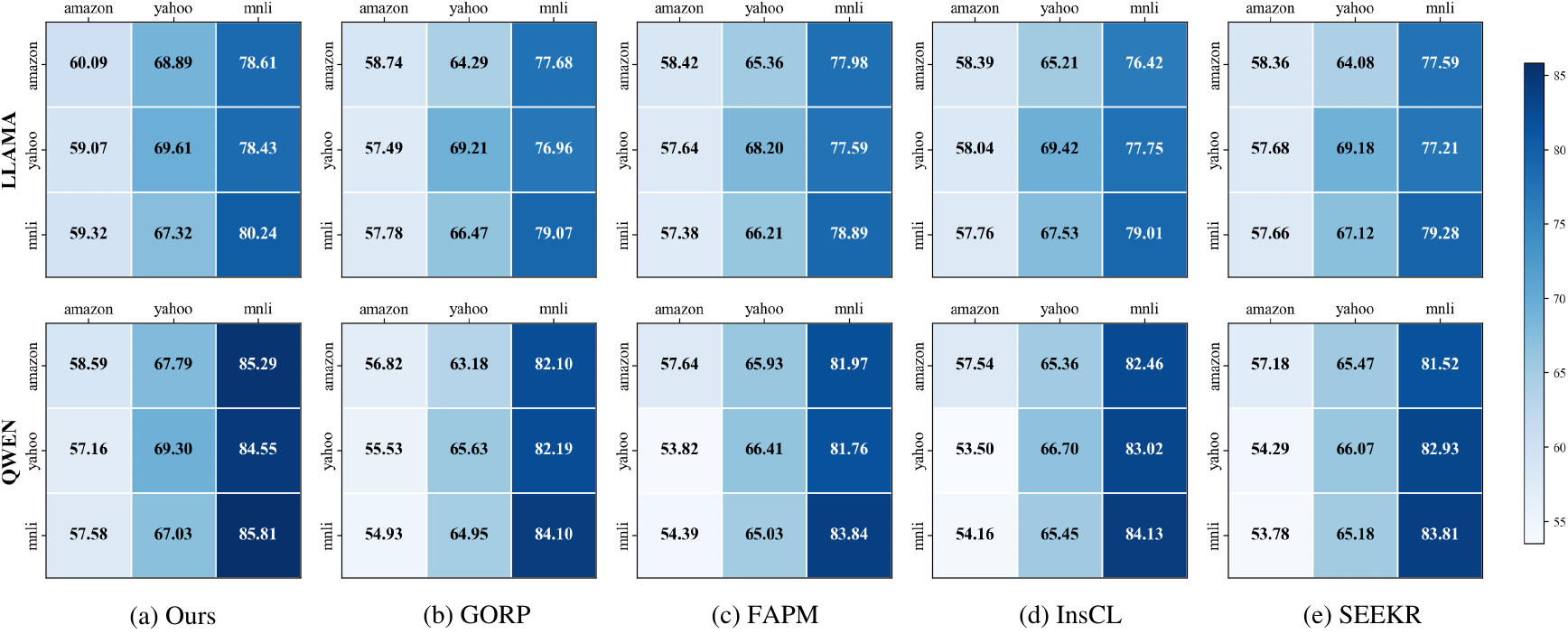}
  \caption{
  Full heatmap analysis of cross-stage generalization under different synthesis methods.
  Rows denote model states after sequentially learning up to the corresponding task in the stream, and columns denote target evaluation datasets.
  For example, the Amazon row represents the model after learning the Yelp $\rightarrow$ Amazon stream, while the MNLI row represents the model after learning up to MNLI.
  Darker cells indicate higher downstream accuracy.
  SynLearner achieves stronger and more consistent performance across both seen and unseen target datasets on LLaMA and Qwen backbones.
  }
  \label{fig:transfer_heatmap_full}
\end{figure*}

\paragraph{Full Heatmap Analysis of Cross-stage Generalization.}

Figure~\ref{fig:transfer_heatmap_full} provides the complete heatmap analysis for all compared methods.
Different from a single-task adaptation matrix, each row represents a model checkpoint obtained after sequentially learning up to a specific stage in the task stream.
For example, the Amazon row corresponds to the model trained through the Yelp $\rightarrow$ Amazon stage, while the MNLI row corresponds to the model trained through the full stream up to MNLI.
Each column indicates the target dataset used for downstream evaluation.
Therefore, each cell measures how well a model state learned from the stream generalizes to a specific target dataset, including both previously seen and not-yet-seen tasks.

Across both LLaMA and Qwen backbones, SynLearner shows stronger and more consistent performance than the baseline methods across different model states and target datasets.
This indicates that the synthesis behavior learned by SynLearner is not restricted to the current training stage, but remains useful when evaluated on other datasets in the stream.
In particular, SynLearner maintains competitive performance on both earlier targets and later targets, suggesting that it can reuse accumulated synthesis experience while preserving generalization to different task distributions.

Compared with baselines such as GORP, FAPM, InsCL, and SEEKR, which mainly focus on sequential adaptation or knowledge retention, SynLearner explicitly optimizes synthesis behavior through feedback signals at both sample and set levels.
The heatmap results therefore provide additional evidence that SynLearner learns reusable synthesis strategies rather than merely fitting the current stage of the stream.

\paragraph{Generalization to Mathematical Reasoning Tasks.}

To further examine whether \textbf{SynLearner} generalizes beyond the original NLP understanding task stream, we conduct supplementary experiments on mathematical reasoning tasks.
Specifically, we append a fifth task after the original stream 
(\textit{Yelp} $\rightarrow$ \textit{Amazon} $\rightarrow$ \textit{Yahoo} $\rightarrow$ \textit{MNLI}) 
and evaluate \textbf{GSM8K}~\citep{gsm8k} and \textbf{MATH-500}~\citep{math500}.
Unlike the original benchmarks, these tasks require structured numerical reasoning and answer derivation, providing a stronger test of cross-task-type generalization.

Due to computational constraints, we adopt a simplified evaluation protocol.
We keep the original Step1--Step4 stream unchanged and compare two representative settings: \textbf{Ori}, where the downstream model is trained without synthetic augmentation, and \textbf{Ours}, where \textbf{SynLearner} is used to generate synthetic data under the same StreamSynth pipeline.
The goal is not to establish state-of-the-art mathematical reasoning performance, but to test whether synthesis experience accumulated from earlier tasks can provide useful transfer to a new reasoning-oriented domain.

As reported in Table~\ref{tab:math_generalization}, SynLearner improves Step 5 performance on both GSM8K and MATH-500 across LLaMA3.1-8B and Qwen2.5-7B.
The improvements are especially notable on MATH-500, suggesting that the learned synthesis behavior is not limited to the original task stream and can provide useful generalization signals for more structured reasoning tasks.
Nevertheless, these results remain supplementary, and broader evaluation on instruction generation, open-domain QA, code synthesis, and agentic tasks is left for future work.

\paragraph{Scaling Behavior across Model Sizes.}

We further report detailed scaling results in Table~\ref{tab:scaling_law}, which complements the analysis in Figure~\ref{fig:order_analyze}(c).
As expected, synthesis performance generally improves as model size increases, particularly when scaling from small models (1B--1.5B) to mid-sized models.

However, the performance gains become marginal beyond the 7B--8B scale.
Models with 13B--14B parameters yield only limited improvements over 7B--8B models across most datasets.
This saturation effect suggests that, under the StreamSynth setting, synthesis performance is no longer primarily constrained by model capacity once a sufficient scale is reached.
Instead, the learning framework and experience accumulation mechanism play a more central role in determining synthesis quality.

\section{Impact and Future Directions}
\label{sec:impact_future}

Our proposed framework, \textbf{SynLearner}, introduces \textbf{StreamSynth} as a new perspective for studying synthetic data generation under evolving task streams.
Rather than treating synthesis as a collection of isolated procedures, this work highlights the potential of learning reusable and transferable synthesis behaviors from experience.
Such a perspective is particularly relevant for real-world scenarios where data requirements change over time and static generation pipelines are insufficient.

\paragraph{Impact.}
This work contributes to the data synthesis literature in the following aspects:

\begin{itemize}[nolistsep, leftmargin=*]
    \item \textbf{A new perspective on synthesis learning.}  
    By formulating \textbf{StreamSynth}, we provide a unified setting for analyzing how synthesis behaviors can be acquired, refined, and reused across tasks.
    This perspective complements existing task-centric synthesis approaches and encourages viewing data generation as an adaptive, experience-driven process.

    \item \textbf{Towards data-efficient and reusable synthesis.}  
    Our results demonstrate that synthesis strategies learned from earlier tasks can benefit future ones, reducing the need for repeated prompt engineering or full retraining.
    This opens the possibility of more data-efficient pipelines for continuously evolving applications.
\end{itemize}

\paragraph{Future Directions.}  
Building upon this foundation, we plan to extend \textbf{StreamSynth} along three concrete directions.

First, we aim to explore \textbf{more domain-rich synthesis streams}, including tasks with stronger structural constraints such as mathematical reasoning, code generation, and multimodal data synthesis. 
These domains introduce more complex validity and consistency requirements, and thus provide a natural testbed for evaluating whether learned synthesis behaviors can generalize beyond classification-style tasks.

Second, we will investigate \textbf{improved self-correction mechanisms during generation}. 
While the current framework relies on external reward signals for post-hoc optimization, incorporating intermediate self-evaluation or revision during synthesis may enable more efficient error correction and reduce reliance on downstream filtering.
Such mechanisms could further enhance stability and reliability in long task streams.

Third, we plan to study \textbf{adaptive reward structures} that dynamically evolve with task progression.
Rather than using fixed reward weights across all stages, future work may adjust reward emphasis based on task difficulty, stream position, or observed synthesis behavior, allowing the learning signal to better align with evolving synthesis objectives.

We hope that \textbf{StreamSynth} can serve not only as a research framework but also as a practical benchmark, encouraging further investigation into continual, reliable, and data-efficient synthesis learning systems.

    

\section{Artifact, Data, and Ethical Considerations}
\label{app:artifact_ethics}

This appendix provides additional information regarding the artifacts, data usage, and ethical considerations of this work, following the ACL responsible research checklist.

\paragraph{Artifact License and Terms of Use.}
Our work does not introduce new datasets; instead, it relies on widely used public benchmarks, including Amazon Reviews, Yelp, Yahoo Answers, and MNLI.
All datasets are used in accordance with their original licenses and terms, which permit research use.
The code and synthetic data generation framework developed in this work will be released for research purposes only, under a permissive academic license.
We do not claim ownership over any original dataset content, and all derived artifacts are intended strictly for non-commercial research use.

\paragraph{Consistency with Intended Use.}
The use of all existing datasets in this work is consistent with their original intended purpose, namely academic research on natural language understanding and generation.
Our framework generates synthetic data derived from these datasets solely to study learning behaviors under the proposed \textbf{StreamSynth} setting.
We explicitly restrict the intended use of generated synthetic data to research contexts, and do not advocate its deployment in real-world applications without additional validation, filtering, and compliance checks.
This is compatible with the access conditions of the original datasets.

\paragraph{Personally Identifying Information and Offensive Content.}
The datasets used in this work are standard NLP benchmarks that have been previously curated and released for research use.
They do not intentionally contain personally identifying information.
Nevertheless, we follow standard preprocessing procedures and do not introduce or amplify sensitive attributes during synthesis.
Since synthetic data generation may potentially reflect biases or undesirable patterns present in source data, all generated samples are used exclusively for controlled experimental evaluation, rather than downstream deployment.

\paragraph{Artifact Documentation and Coverage.}
We provide documentation for the released artifacts, including descriptions of the task streams, dataset domains, and experimental configurations used in \textbf{StreamSynth}.
The tasks primarily cover English-language text classification and natural language inference, focusing on sentiment analysis, topic classification, and reasoning.
We do not make claims about demographic representativeness beyond what is provided by the original datasets.
The documentation aims to clearly state the scope and limitations of the artifacts to facilitate reproducibility and responsible reuse.

\section{The Use of Large Language Models}
\label{llm_use}
We disclose that LLMs were employed solely for translation and language refinement purposes. All research ideas, experimental design, implementation, analysis, and conclusions are the sole responsibility of the authors. We have carefully verified the accuracy and integrity of the manuscript to ensure that no false or misleading content was introduced by the use of LLMs.

\section{Algorithmic Details of \textbf{SynLearner}}
\label{appendix:algorithm}

This section presents the complete algorithmic formulation of the proposed \textbf{SynLearner} framework and its \textbf{Hierarchical Reward Optimization (HRO)} strategy.
Algorithm~\ref{alg:streamsynth} outlines the full learning procedure under the \textbf{StreamSynth} setting, where the model incrementally acquires synthesis capability across a sequence of tasks.
Algorithm~\ref{alg:reward} further specifies the hierarchical reward construction, which integrates sample-level and set-level feedback to jointly regulate synthesis quality, diversity, and task consistency.

\subsection{SynLearner Training Procedure under StreamSynth}

Algorithm~\ref{alg:streamsynth} illustrates the overall training loop of \textbf{SynLearner} over a task stream $\mathcal{T} = \{\mathcal{T}_1, \dots, \mathcal{T}_K\}$.
Tasks are processed sequentially, and no task-specific parameters are retained once the stream advances.

For each incoming task, SynLearner first applies \textbf{EFT} to obtain a task-adapted initialization, which stabilizes early-stage synthesis and facilitates efficient adaptation.
Based on this initialization, the model then performs reinforcement optimization guided by \textbf{HRO}, where both sample-level rewards and set-level rewards are computed on synthesized data batches.
These hierarchical signals jointly encourage instance-level correctness while maintaining global diversity and distributional coverage.

Crucially, both the optimized model parameters and the synthesized data are carried forward to subsequent tasks.
This progressive transfer enables the model to accumulate synthesis experience over the task stream, allowing later tasks to benefit from previously acquired synthesis behaviors without explicit replay or task-specific memory.


\begin{algorithm*}[t]
\caption{SynLearner Framework for StreamSynth with Hierarchical Reward Optimization}
\label{alg:streamsynth}
\begin{algorithmic}[1]
\Require Streaming task sequence $\mathcal{T} = \{T_1, T_2, \dots, T_n\}$, base synthesis model $M$, maximum RL steps $K$, prompt pool $\mathcal{P}$, embedding model $\mathcal{E}$, trade-off coefficient $\lambda$, weights $\gamma_1,\gamma_2,\gamma_3$, hyperparameter $k$
\Ensure Trained synthesis model $M^*$
\State Initialize model $\mathcal{M}_0 \leftarrow M$
\For{each task $T_t \in \mathcal{T}$}
    \State Select task-adaptive prompt $p_t \in \mathcal{P}$
    \State Generate candidate samples for task $T_t$
    \State Collect task-specific synthesis data $\mathcal{D}^{(t)}_{\text{sft}}$
    \State $\mathcal{M}_t \leftarrow \text{EFT}(\mathcal{M}_{t-1}, \mathcal{D}^{(t)}_{\text{sft}})$
    \For{$step = 1$ to $K$}
        \State Sample input $x^{(t)} \sim T_t$, generate output $y^{(t)} \sim \mathcal{M}_t(x^{(t)})$
        \State Compute sample-level reward $r^{(t)}_{\text{sample}} \leftarrow RS_t(y^{(t)})$
        \State Collect mini-batch $\mathcal{B}_t$
        \State Compute set-level reward $r^{(t)}_{\text{set}} \leftarrow DS_t(y^{(t)} \mid \mathcal{B}_t)$
        \State $r^{(t)}_{\text{total}} \leftarrow \lambda r^{(t)}_{\text{sample}} + (1-\lambda) r^{(t)}_{\text{set}}$
        \State $\mathcal{M}_t \leftarrow \text{RL-Update}(\mathcal{M}_t, x^{(t)}, y^{(t)}, r^{(t)}_{\text{total}})$
        \State $M^* \leftarrow \mathcal{M}_t$
    \EndFor
    \State Save updated model for transfer to next task
\EndFor
\State \Return $M^*$
\end{algorithmic}
\end{algorithm*}

\subsection{Hierarchical Reward Optimization Mechanism}
Algorithm~\ref{alg:reward} details the \textbf{Hierarchical Reward Optimization (HRO)} process used during reinforcement learning.  
The total reward is composed of two parts:
(1) a \textbf{sample-level reward} that measures the structural, fluency, and task relevance quality of each generated instance, and  
(2) a \textbf{set-level reward} that encourages global diversity within the generated batch.  
This hierarchical design allows SynLearner to balance between local accuracy and global diversity, guiding the synthesis process toward both stable and diverse sample generation.

\begin{algorithm*}[t]
\caption{Hierarchical Reward Optimization (Sample-level + Set-level)}
\label{alg:reward}
\begin{algorithmic}[1]
\Require Generated sample $y^{(t)}$, mini-batch $\mathcal{B}_t = \{x^{(t)}_1,\dots,x^{(t)}_m\}$, embedding model $\mathcal{E}$, coefficients $\lambda, \gamma_1, \gamma_2, \gamma_3$, hyperparameter $k$
\Ensure Final reward $r^{(t)}_{\text{total}}$
\State Evaluate structural validity $S_{\text{struct}}(y^{(t)})$
\State Evaluate fluency and naturalness $S_{\text{fluent}}(y^{(t)})$
\State Evaluate task-specific relevance $S^{(t)}_{\text{rel}}(y^{(t)})$
\State Aggregate sample-level reward:
\begin{equation*}
RS_t(y^{(t)}) = \gamma_1 S_{\text{struct}}(y^{(t)})
+ \gamma_2 S_{\text{fluent}}(y^{(t)})
+ \gamma_3 S^{(t)}_{\text{rel}}(y^{(t)})
\end{equation*}
\State Obtain embedding $e^{(t)} \leftarrow \mathcal{E}(y^{(t)})$
\For{each $x^{(t)}_j \in \mathcal{B}_t$}
    \State Compute cosine similarity $s_j = \cos\big(e^{(t)}, \mathcal{E}(x^{(t)}_j)\big)$
\EndFor
\State Normalize proximity weights:
\begin{equation*}
w^{(t)}_j = \frac{s_j}{\sum_{k=1}^{m} s_k}
\end{equation*}
\State Compute local density:
\begin{equation*}
D_t(y^{(t)}) = \sum_{j=1}^{m} w^{(t)}_j \cdot s_j
\end{equation*}
\State Convert density to distinctiveness score:
\begin{equation*}
DS_t(y^{(t)}) = \exp\big(-k \cdot D_t(y^{(t)})\big)
\end{equation*}
\State Compute final reward:
\begin{equation*}
r^{(t)}_{\text{total}} = \lambda RS_t(y^{(t)}) + (1-\lambda) DS_t(y^{(t)})
\end{equation*}
\State \Return $r^{(t)}_{\text{total}}$
\end{algorithmic}
\end{algorithm*}

\end{document}